\documentclass{article}

\usepackage{microtype}
\usepackage{graphicx}
\usepackage{subfigure}
\usepackage{booktabs} %
\usepackage{makecell}
\usepackage{multirow}
\usepackage{soul}

\usepackage{hyperref}

\usepackage[accepted]{icml2024}

\usepackage{amsmath}
\usepackage{amssymb}
\usepackage{mathtools}
\usepackage{amsthm}
\newcommand{\modelname}{MoE-Mamba}

\newcommand{\kkm}[1]{} %

\newcommand{\ff}{feed-forward }
\newcommand{\FF}{Feed-Forward }
\newcommand{\Nexperts}{N_{\text{experts}}}
\newcommand{\dmodel}{d_\text{model}}
\DeclareMathOperator*{\argmax}{arg\!\max}

\newcommand{\smalll}{$\Box_{25\text{M}}$}

\newcommand{\largee}{$\Box_{100\text{M}}$}

\newcommand{\transformerSmall}{$\text{Transformer}_{25\text{M}}$}
\newcommand{\moeSmall}{$\text{Transformer-MoE}_{25\text{M}}$}
\newcommand{\mambaSmall}{$\text{Mamba}_{25\text{M}}$}

\newcommand{\moemambaSmall}{$\text{\modelname}_{25\text{M}}$}

\newcommand{\moeLarge}{$\text{Transformer-MoE}_{100\text{M}}$}
\newcommand{\mambaLarge}{$\text{Mamba}_{100\text{M}}$}
\newcommand{\moemambaLarge}{$\text{\modelname}_{100\text{M}}$}

\usepackage[capitalize,noabbrev]{cleveref}

\theoremstyle{plain}

\theoremstyle{definition}

\theoremstyle{remark}

\usepackage[textsize=tiny]{todonotes}

\icmltitlerunning{MoE-Mamba: Efficient Selective State Space Models with Mixture of Experts}

\begin{document}

\onecolumn %
\icmltitle{MoE-Mamba: Efficient Selective State Space Models with Mixture of Experts}

\icmlsetsymbol{equal}{*}

\begin{icmlauthorlist}
\icmlauthor{Maciej Pióro}{ideas,paos}
\icmlauthor{Kamil Ciebiera}{ideas,uw}
\icmlauthor{Krystian Król}{ideas,uw}
\icmlauthor{Jan Ludziejewski}{ideas,uw}
\icmlauthor{Michał Krutul}{ideas,uw}
\icmlauthor{Jakub Krajewski}{ideas,uw}
\icmlauthor{Szymon Antoniak}{}
\icmlauthor{Piotr Miłoś}{ideas,impan,uw}
\icmlauthor{Marek Cygan}{uw}
\icmlauthor{Sebastian Jaszczur}{ideas,uw}
\end{icmlauthorlist}

\icmlaffiliation{ideas}{IDEAS NCBR}
\icmlaffiliation{uw}{University of Warsaw}
\icmlaffiliation{paos}{Polish Academy of Sciences}
\icmlaffiliation{impan}{Institute of Mathematics, Polish Academy of Sciences}

\icmlcorrespondingauthor{Sebastian Jaszczur}{s.jaszczur@uw.edu.pl}

\icmlkeywords{Machine Learning, Mamba, Mixture of Experts, MoE, Transformer, LLM}

\vskip 0.3in

\printAffiliationsAndNotice{Detailed authors' contributions are listed in Appendix \ref{app:contributions}.}

\begin{abstract}
State Space Models (SSMs) have become serious contenders in the field of sequential modeling, challenging the dominance of Transformers. At the same time, Mixture of Experts (MoE) has significantly improved \mbox{Transformer-based} Large Language Models, including recent state-of-the-art open models. We propose that to unlock the potential of SSMs for scaling, they should be combined with MoE. We showcase this on Mamba, a recent SSM-based model that achieves remarkable performance. Our model, \modelname{}, outperforms both Mamba and baseline Transformer-MoE. In particular, \modelname{} reaches the same performance as Mamba in \emph{$2.35\times$ fewer training steps} while preserving the inference performance gains of Mamba against Transformer. 
\end{abstract}

\begin{figure*}[bht]
\centering
\includegraphics[width=.9\textwidth]{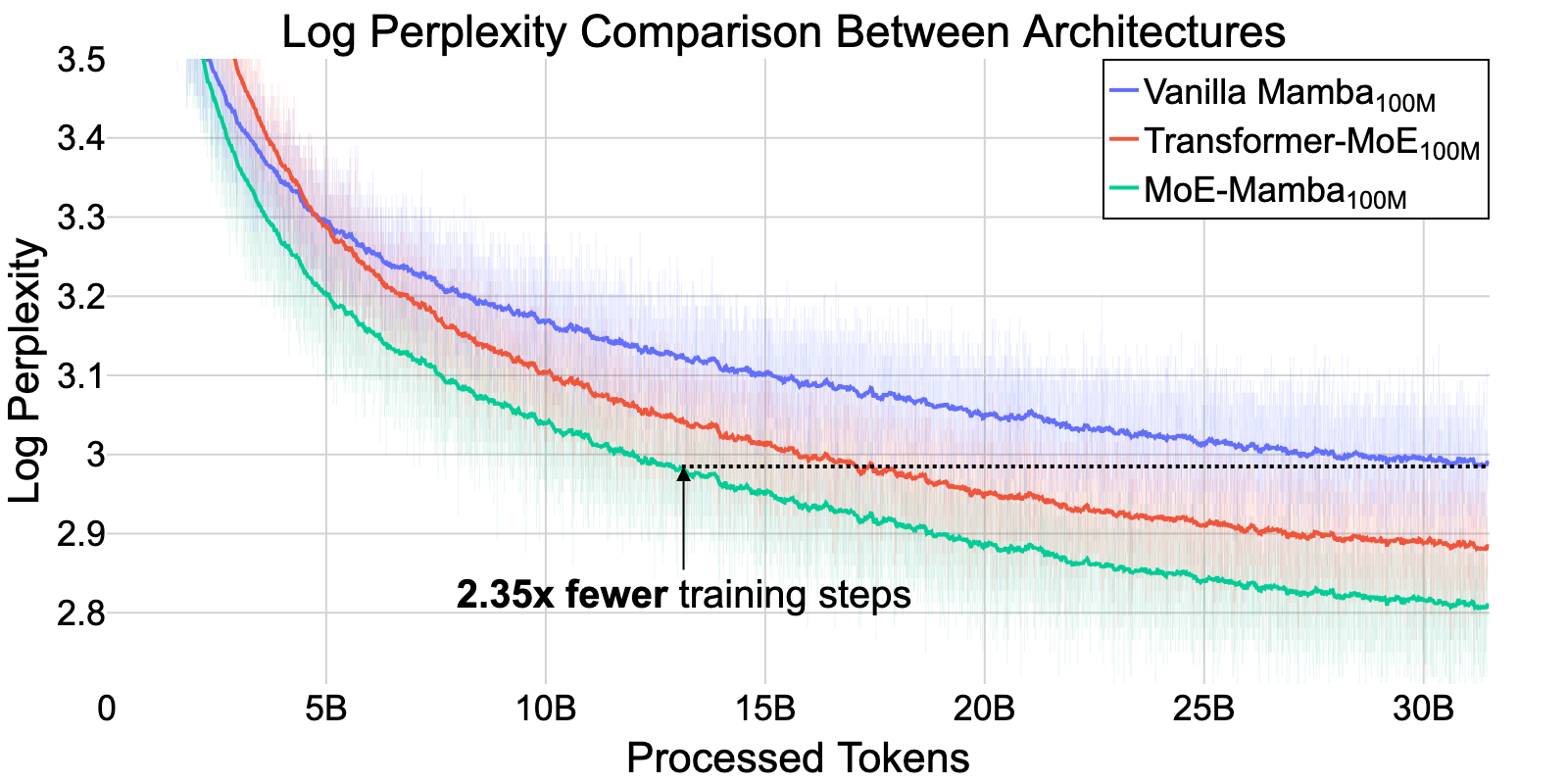} %

\caption{Log perplexity throughout the training. From top to bottom: \mambaLarge; \moeLarge; \moemambaLarge.}
\label{fig:figure1}
\end{figure*}

\section{Introduction} \label{intro}

Large Language Models (LLMs) have emerged as a cornerstone in the ongoing AI revolution \citep{brown2020language,chowdhery2023palm, lewkowycz2022solving, openai2023gpt4, geminiteam2023gemini}. Their remarkable effectiveness is primarily attributed to the Transformer architecture \citep{vaswani2017attention} and training on an internet-wide scale, e.g., \citep{together2023redpajama}. Yet, questions remain: Should Transformers be the only architecture used for LLMs? Can we scale language models even further, and if so, how can this be achieved?

\looseness=-1 Regarding the first question, State Space Models (SSMs), e.g., \cite{gu2022efficiently, gu2021combining, gu2022parameterization, gupta2022diagonal, li2022makes, ma2022mega, orvieto2023resurrecting, smith2023simplified}, have been increasingly gaining attention. This recognition is due to their capability for linear-time inference, highly parallelized training, and strong performance in tasks requiring long-context processing, such as those illustrated by the Long Range Arena \citep{tay2020long}. Notably, a recent addition to this category, Mamba \cite{gu2023mamba}, has shown impressive results through its selective mechanism and hardware-aware design, positioning it as a promising contender to the attention-based Transformer architecture.

Scaling is believed to be a critical factor in developing powerful AI systems \citep{sutton2019bitter}. The Mixture of Experts (MoE) approach \cite{moe1991}, a set of techniques that enables an increase in model parameters with minimal impact on computational demands, plays a significant role. Due to their sparse activation, MoEs can be efficiently scaled up to trillions of parameters, as demonstrated by \citet{shazeer2017outrageously, fedus2022switch}. MoE variants \cite{fedus2022switch, du2022glam} are now routinely used in LLMs, as exemplified in the recent Mixtral model \cite{jiang2024mixtral}.

In this paper, \emph{we advocate that to unlock the potential of SSMs for scaling up, they should be combined with Mixture of Experts (MoE)}. To this end, we introduce \textbf{\modelname{}}, a model that combines Mamba \cite{gu2023mamba} with a Switch layer \cite{fedus2022switch}. \modelname{} enables efficiency gains of both SSMs and MoE, outperforming Mamba and Transformer-MoE, see Figure \ref{fig:figure1}. Through comprehensive studies, we confirm that the effect is robust to the design choices and the number of experts. Our results indicate a very promising research direction that may allow scaling SSMs beyond tens of billions of parameters and compete with the largest SoTA language models. 

In summary, our contributions are as follows:
\begin{itemize}
    \item We introduce \modelname{}, a model that combines Mamba with a Mixture of Experts layer. \modelname{} enables efficiency gains of both SSMs and MoE while reaching the same performance as Mamba in $2.35\times$ fewer training steps.
    \item Via comprehensive studies, we confirm that the improvement achieved by \modelname{} is robust to varying model sizes, design choices, and the number of experts. 
    \item We explore and compare multiple alternative methods of integrating Mixture of Experts within the Mamba block.
\end{itemize}

\section{Related Work}

\looseness=-1\paragraph{State Space Models and Related Attention-Free Architectures }
 State Space Models (SSMs) \cite{gu2022efficiently, gu2021combining, gu2022parameterization, gupta2022diagonal, li2022makes, ma2022mega, orvieto2023resurrecting, smith2023simplified} form a family of architectures used for sequence modeling. Stemming from signal processing, these models can be seen as a combination of RNNs and CNNs \cite{gu2023mamba}. Although they potentially offer considerable benefits, a number of issues have been identified with SSMs \cite{gu2022efficiently}, preventing SSMs from becoming the leading architecture in the task of language modeling. However, recent breakthroughs \cite{gu2022efficiently,fu2023hungry,smith2023simplified, gu2023mamba}, have allowed deep SSMs to be increasingly competitive against Transformers \citep{vaswani2017attention}. In particular, Mamba \cite{gu2023mamba}, studied in this paper, has shown impressive results through its selective mechanism and hardware-aware design, which allows scaling to billions of parameters while retaining computational efficiency and strong performance. Besides SSMs, numerous other architectures have been proposed that do not rely on the quadratic attention mechanism \cite{zhai2021attention,poli2023hyena,sun2023retentive, peng2023rwkv}.

\paragraph{Mixture of Experts}
Mixture of Experts (MoE) is a class of techniques that allow drastically increasing the number of parameters of a model without much impact on the FLOPs required for the model's training and inference. Introduced by \citet{moe1991,716791}, MoE was applied in the context of NLP by \citet{shazeer2017outrageously}. MoE models benefit from sparse activation - for each token processed, only a subset of the model's parameters is used. Due to their computational demands, \ff layers in Transformers have become the standard target of various MoE techniques \cite{lepikhin2020gshard,fedus2022switch,du2022glam,zoph2022stmoe}. Scaling properties and generalization abilities of MoE Transformers have been studied more closely by \citet{artetxe2021efficient,clark2022unified, krajewski2024scaling}.

\looseness=-1 More recently, MoE models have found their way onto the open scene \cite{openmoe2023, jiang2024mixtral}. In particular, the Mixtral 8$\times$7B model \cite{jiang2024mixtral} fares comparably to Llama 2 $70$B \cite{touvron2023llama} while requiring only around $1/6$ of its inference computational budget.

Subsequently to the first version of our work, \citet{anthony2024blackmamba} presented an architecture similar to \modelname, showcasing the potential of connecting Mamba with MoE on downstream tasks, which validates our findings. In contrast to their work, we run extensive ablations on the model architecture, number of experts, and other design choices. We also investigate the potential of integrating conditional computation into the Mamba block.

\section{\modelname} \label{sec:architecture}
In this section, we present the architecture details of our model, \modelname{}, see Figure \ref{fig:designs_alt}. We start with a brief overview of the Mamba architecture, followed by a description of the MoE layer. Our main architecture is presented in Section \ref{sec:architectures}, while sections \ref{sec:parallel} and \ref{sec:modifying_block} explore its variants and related approaches.

\subsection{Preliminaries}

\paragraph{Mamba} %
Mamba \citep{gu2023mamba} is a recently proposed SSM-based model that achieves remarkable, Transformer-like performance. By employing a work-efficient parallel scan, Mamba mitigates the impact of the sequential nature of recurrence, whereas fusing GPU operations removes the requirement to materialize the expanded state. Intermediate states necessary for backpropagation are not saved but recomputed during the backward pass, thus reducing memory requirements. The advantages of Mamba over the attention mechanism are especially prominent during inference, as not only is the computational complexity lowered, but also the memory usage is not dependent on the context length. Figure \ref{fig:other_approaches} shows the inner structure of a Mamba layer.

\paragraph{MoE Layer}
In our work, we follow the well-established \cite{zhao2023survey,sanseviero2023moe} and easy-to-implement Switch Transformer MoE design \cite{fedus2022switch} and leave consideration of other MoE designs for future work. 

We assume $\Nexperts$ experts $\{E_i\}_{i=1}^{\Nexperts}$, each being a trainable feed-forward network with the same number of parameters. For each token embedding $x$, we calculate scores $h(x) = W x \in \mathbb{R}^{\Nexperts}$, where $W$ is a trainable linear projection. These are normalized using softmax:
$$p_{i}(x) = \frac{\exp\left(h(x)_i\right)}{\sum_{i=1}^{\Nexperts}\exp\left(h(x)_i\right)}.$$ 
Prior to Switch, top-$k$ routing selecting $k>1$ most suitable experts for each token was deemed necessary. However, Switch successfully simplifies previous MoE approaches by setting $k=1$. Namely, the output of the MoE layer for $x$ is given by:
$$y = p_{I}E_{I}(x),$$
where $I=\argmax_{i} p_i(x)$. 

During batched execution, e.g., in training, each batch contains $N$ tokens. Following the standard procedure, in a case where the assignment of tokens to the experts is not perfect, i.e., some expert $E_f$ is selected by more than $N / \Nexperts$ tokens in the current batch, the excess tokens are dropped and not updated ($\text{capacity factor}=1$). To further encourage an even distribution of tokens to experts, load balancing loss as described by \citet{fedus2022switch} with weight $\alpha=0.01$ is added to the training objective. %

\subsection{\modelname{} Architecture} \label{sec:architectures}

\begin{figure*}[]
    \centering
    \includegraphics[width=1\textwidth]{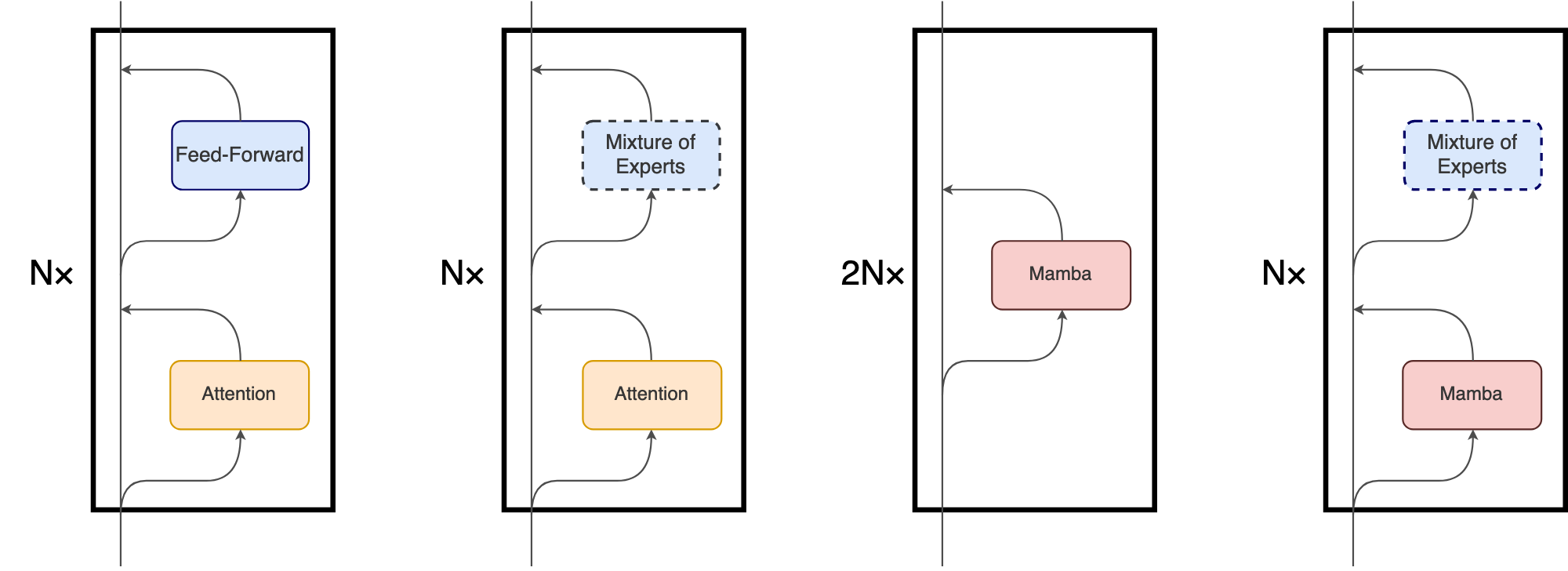}
    \caption{
    Diagrams of the architectures. From the left: vanilla Transformer, Transformer-MoE, Mamba, \modelname{}. 
    }
    \label{fig:designs_alt}
\end{figure*}

The vanilla Mamba architecture consists of multiple Mamba blocks stacked one after another, with each layer's output being added to the residual stream; see Figure \ref{fig:designs_alt}. 
In \modelname{}, we interleave Mamba layers with MoE layers (see Figure \ref{fig:designs_alt}). Note that the vanilla Mamba does not use \ff layers. %

In this way, \modelname{} separates unconditional processing of every token by the Mamba layer - which can efficiently integrate the whole sequence context into an internal representation - and conditional processing by an MoE layer that can apply the most relevant expert (and thus the subset of parameters) for each token. The idea of interleaving conditional and unconditional processing is used in some MoE-based models, typically by alternating vanilla and MoE \ff layers \cite{lepikhin2020gshard,fedus2022switch}.

\subsection{Parallel \modelname{}} \label{sec:parallel}
\begin{figure*}[t]
    \centering
    \includegraphics[width=0.8\textwidth]{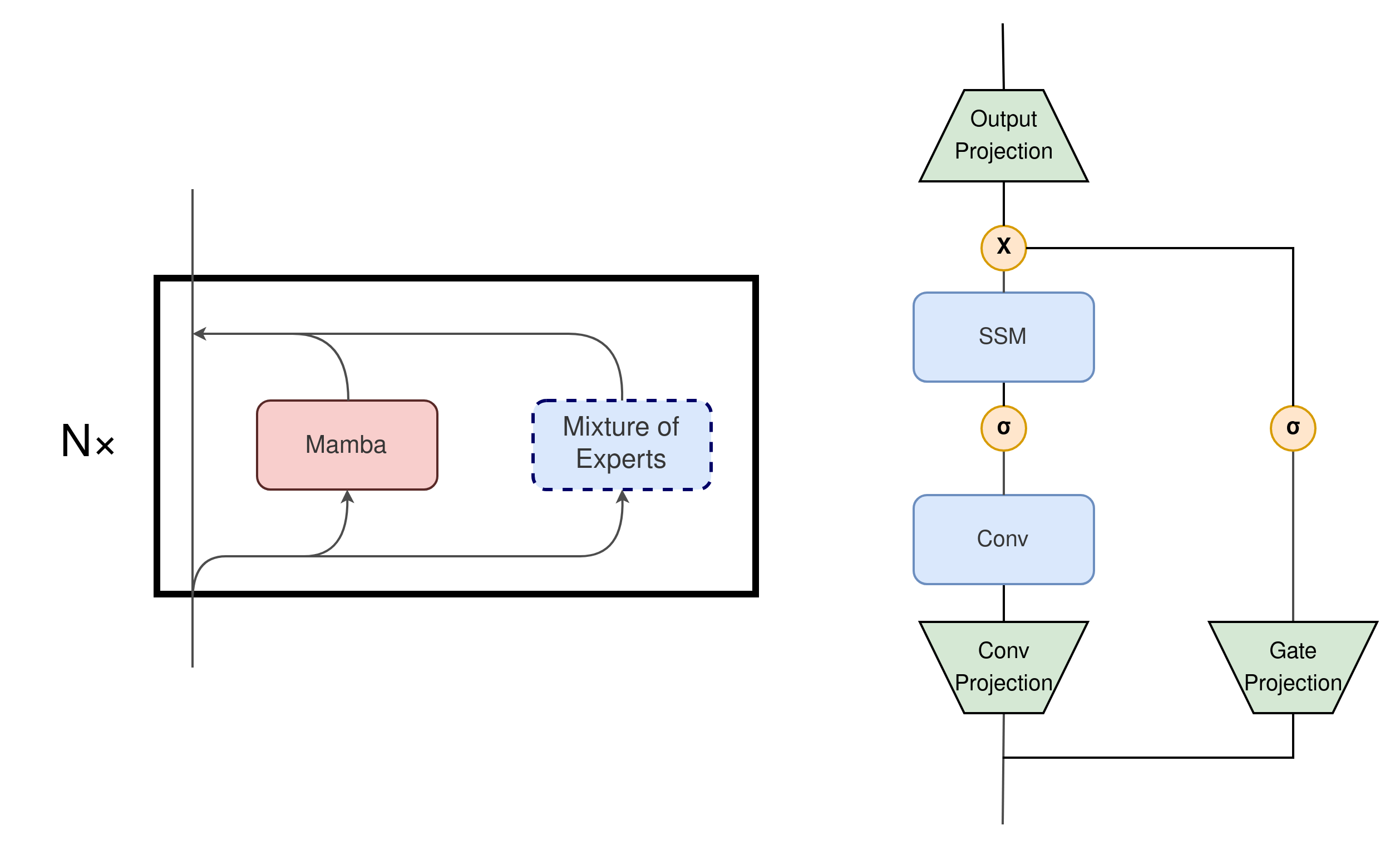}
    \caption{
    Diagram of Parallel \modelname{} architecture (left) and Mamba Block (right). The outputs of the Gate and Conv Projections are $E$ (expansion factor) times bigger than the input, i.e., Conv and SSM operate on vectors $\in \mathbb{R}^{E \cdot \dmodel}$. Vanilla Mamba assumes $E=2$ \cite{gu2023mamba}.
    Expansion factor $E$ determines how much the input vector is scaled up by Gate and Conv Projection and then scaled down by Output Projection, and because of that, it is also proportional to the number of FLOPs and parameters in the Mamba layer.  %
    }
    \label{fig:other_approaches}
\end{figure*}

Apart from interleaving MoE layers with Mamba layers, we explore another design, inspired by \citet{mesh-transformer-jax} and \citet{chowdhery2023palm} in which MoE layer is executed in parallel with Mamba (see Figure \ref{fig:other_approaches}). It achieves positive results, albeit worse than \modelname.

\subsection{Modifying Mamba Block} \label{sec:modifying_block}
In addition to attaching a separate MoE layer to Mamba, we also conducted other experiments, modifying the original block design by \citet{gu2023mamba} to feature conditional MoE computation. Some of the designs show improvements over the baseline architecture and suggest promising future research directions.

\section{Experiments}

In this section we provide empirical validation of our hypothesis that interleaving Mamba with MoE can improve the performance of Mamba. Our main result, see Figure \ref{fig:figure1}, shows that \modelname{} needs \emph{$2.35\times$ fewer training steps} to reach the same performance as Mamba. We also provide a detailed analysis of our design choices. 

\begin{table*}[t]
    \centering
    \caption{Comparison between different architectures. The \smalll{} models were trained on ca. 10B tokens and the \largee{} models were trained on ca. 30B tokens. Note that the parameter counts exclude embedding and output (unembedding) layers (for further discussion of reporting either non-embedding or all parameters, see Appendix \ref{app:params}).
    The numbers of total and active parameters are not matched exactly between similarly sized models due to, among other reasons, the MoE models including routers and Mamba layer not containing precisely $6\dmodel^2$ parameters - a design choice we did not want to modify. We consider those differences to be too small to be significant for our results.} \label{tab:results}
    \begin{tabular}{c c c|c c}
        \toprule
        \textbf{Model} & \textbf{\# Parameters} & \thead{\textbf{\# Active Parameters} \\ \textbf{per Token}} &  \thead{\textbf{Final Log} \\ \textbf{Perplexity}} &  \thead{\textbf{Speedup Over} \\ \textbf{Vanilla Mamba} \\ \textbf{(Training Steps)}} \\
        \midrule
        \mambaSmall       & 27M  & 27M   & 3.34 & 1 \\
        \moemambaSmall{} (ours)    & 542M & 26M   & \textbf{3.19} &  \textbf{1.76} \\
        \moeSmall         & 545M & 25M   & 3.23 &  1.56 \\
        \transformerSmall & 25M  & 25M   & 3.43 & $>$1 \\
        \midrule
        \mambaLarge       & 121M  & 121M & 2.99 & 1 \\
        \moemambaLarge{} (ours)    & 2439M & 117M & \textbf{2.81} & \textbf{2.35} \\
        \moeLarge         & 2454M & 114M & 2.88 & 1.79 \\
        \bottomrule
    \end{tabular}
\end{table*}

\subsection{Training Setup}

We compare \modelname{} to three baselines: Mamba, Transformer, and Transformer-MoE. All models in our experiments are decoder-only.

In the standard Transformer architecture, a single attention layer contains $4\dmodel^2$ parameters, whereas a \ff layer contains $8\dmodel^2$ parameters. A single Mamba layer contains slightly over $6\dmodel^2$ \cite{gu2023mamba} parameters.
To be able to compare \modelname{} to Transformer-based and Mamba baselines, we scale down the size of each expert in our model (we set $d_\text{expert}=3\dmodel$). This way, we can keep both the number of blocks and the number of active parameters per token roughly the same in all models of similar size. Active parameters denote those used to calculate the output for a given token (e.g., typically, only one expert in each MoE layer is active). For a discussion of the relation of active parameters and FLOPs, see Appendix \ref{app:active_params}.

Due to computational constraints, we perform most of our experiments on smaller, \smalll{} models and validate our findings on \largee{} models.

\looseness=-1 We train the models on C4 dataset \cite{raffel2020exploring} on the next token prediction task using cross entropy as the loss function. 
We use EMA-smoothed ($\alpha=0.001$) training log perplexity as the comparison metric for both final loss and speedup measurements as it is a more fine-grained comparison metric than test log perplexity. The test log perplexity comparison for \largee{} models can be found in Appendix \ref{app:test_loss}.
All models use the GPT2 tokenizer \cite{radford2019language}.  We tune the learning rate separately for all \smalll{} models and reuse it when training their \largee{} counterparts. When training \moeLarge, we divide the learning rate by two due to repeated instabilities. See Appendix \ref{app:hyperparams} for further details and hyperparameters.
The main experiments, described in section \ref{sec:main_results}, use around 10B tokens for \smalll{} models and around 30B tokens for \largee{} models. The experiments described in further sections use 1B tokens.

\begin{figure*}[ht]
\centering
\includegraphics[width=1\textwidth]{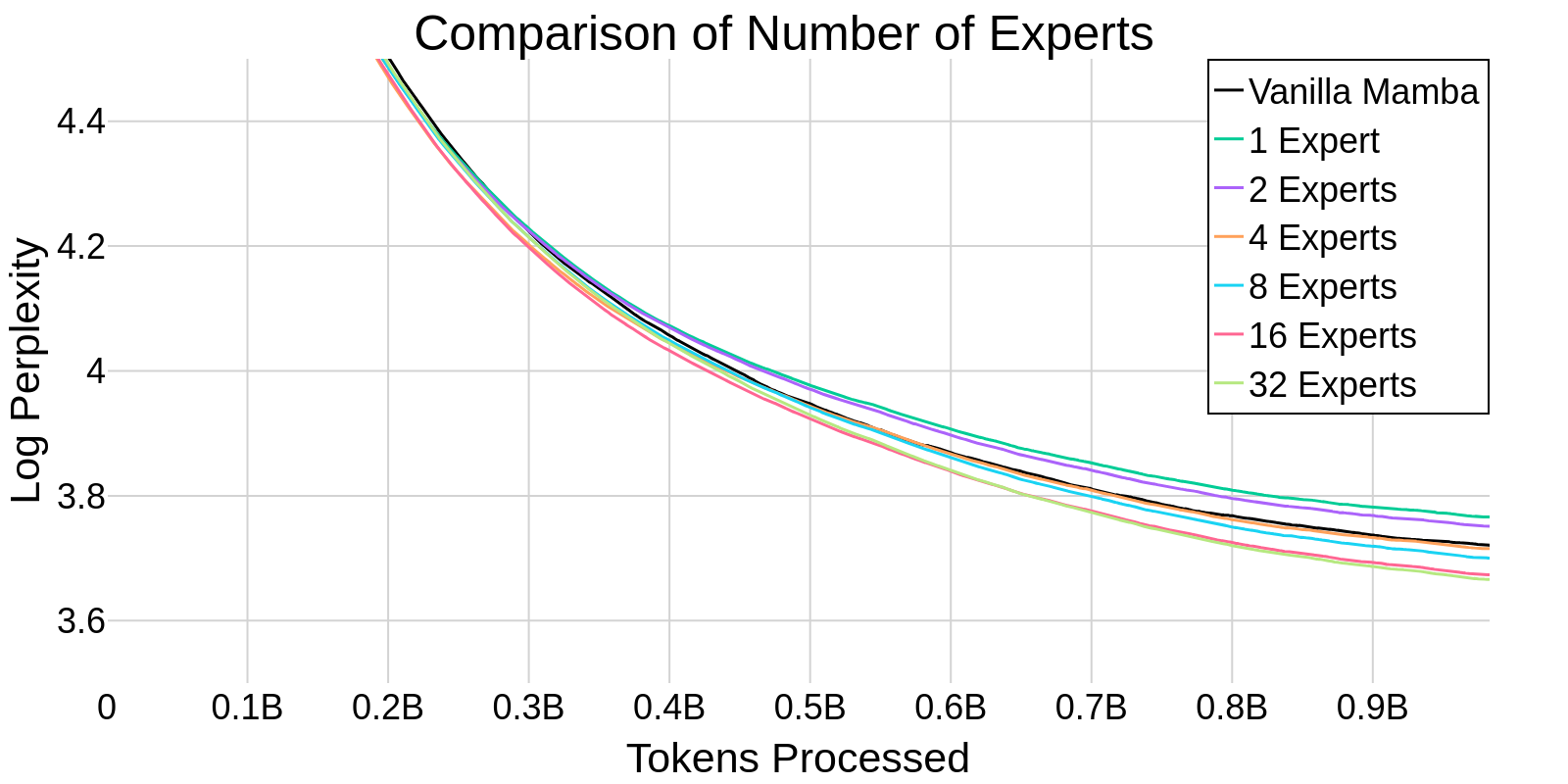}
\caption{Smoothed training loss (log perplexity) for a differing number of experts for \modelname{} with ca. 26M active non-embedding parameters. The final log perplexity improves monotonically as the number of experts increases.
}
\label{fig:experts_comparison}
\end{figure*}

\vspace{-0.1cm}
\subsection{Main Results} 
\vspace{-0.1cm}
\label{sec:main_results}

Table \ref{tab:results} presents the comparison between training results of \modelname{} and baselines; see also Figure \ref{fig:figure1} for log perplexity curves. 
\modelname{} shows a remarkable improvement over the vanilla Mamba model across both model sizes. 
Notably, \moemambaLarge{} was able to
perform on par with
vanilla \mambaLarge{} with $2.35\times$ speedup in terms of processed tokens.
For~\smalll{} model size, those performance gains are lower, probably due to a lower number of training tokens. More generally, we observe that the gains increase over the training, oscillating around $1.6\times-\ 1.9\times$ for \smalll{} models after the initial training period. Further discussion of the speedup can be found in Appendix~\ref{app:speedup}. %
We observe that \modelname{} performs better than the corresponding Transformer-MoE, which strengthens the findings by \citet{gu2023mamba} that Mamba is a competitive alternative to the Transformer.

\subsection{Optimal Ratio of Active Parameters in Mamba and MoE}  \label{sec:optimal_ratio}
In this section, we investigate the optimal ratio of active parameters in the Mamba layer to active parameters in the MoE layer while keeping the total number of parameters fixed. Under these constraints, a given ratio determines the so-called expansion factor $E$ of the Mamba layer, the number of experts, and their size as detailed in Table \ref{tab:various_ratios} (see also Figure \ref{fig:other_approaches} for Mamba design).

\begin{figure}[ht] %
\centering
\includegraphics[width=0.5\textwidth]{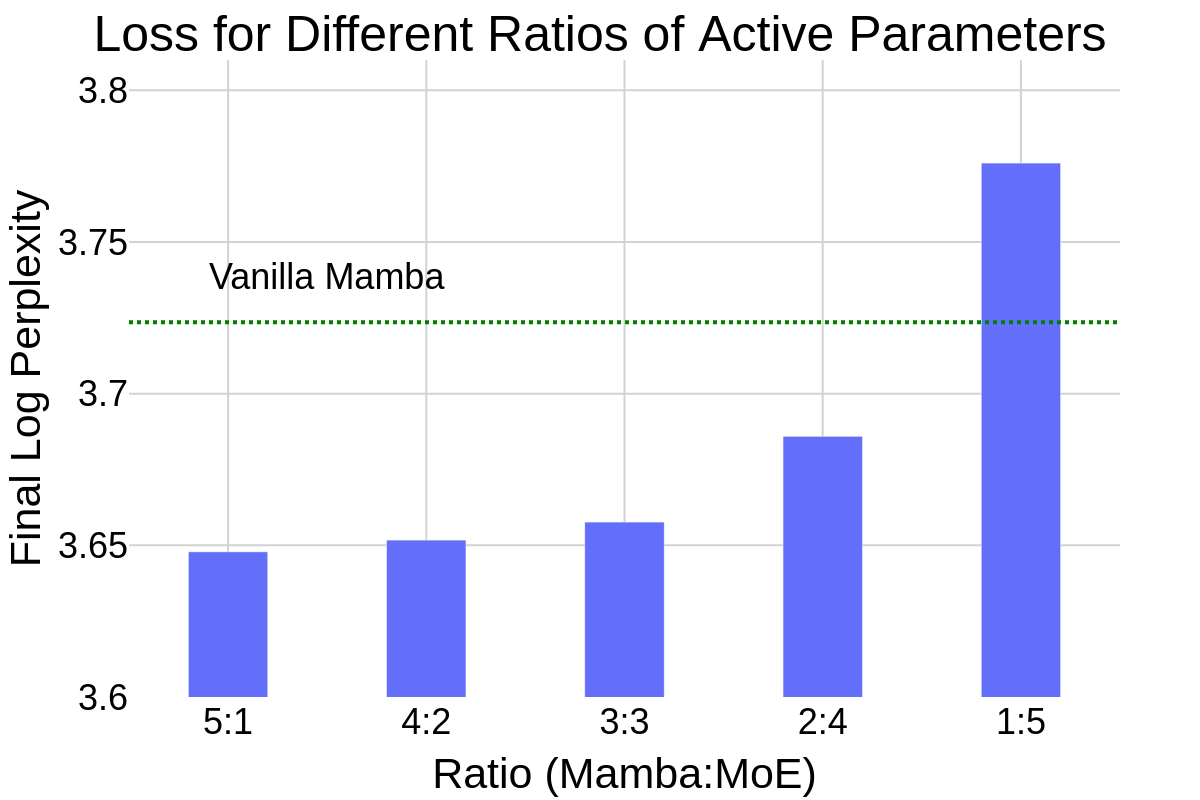} %
\caption{Final log perplexity at different ratios of active Mamba-to-MoE active parameters. Note that MoE contains the majority of the total parameters in each model. For further discussion of the ratios explored, see Appendix \ref{app:optimal_ratio}. }
\label{fig:optimal_ratio}
\end{figure}

The results are presented in Figure \ref{fig:optimal_ratio}. We observe that increasing the number of active Mamba parameters improves the performance. However, the gains become marginal after reaching the $3:3$ ratio, and higher ratios are impractical due to inefficient hardware utilization and high routing costs caused by a large number of experts. We default to this choice in all other experiments.

\begin{table}[]
    \centering
    \caption{Comparison of different ratios of parameters between Mamba and MoE. The $E=2$ corresponds to \moemambaSmall. The total number of parameters in all models is 542M and the number of active parameters per token is 26M.}
    \begin{tabular}{c|c|c|c}
        \toprule
          \thead{ Ratio \\ $N^{\text{act. params}}_{\text{Mamba}} : N^{\text{act. params}}_{\text{MoE}}$} & \thead{Expansion \\ Factor \\ $E$ (Mamba)}  & \thead{ Expert \\ Size  }  & \thead{Number \\ of \\ Experts} \\ \midrule
        $1:5$ & \small $\frac{2}{3}$  & 2560 & 19 \\ 
        $2:4$ & \small $1\frac{2}{3}$ & 2048 & 24 \\ 
        $3:3$ & \small $2$            & 1536 & 32 \\ 
        $4:2$ & \small $2\frac{2}{3}$ & 1024 & 48 \\ 
        $5:1$ & \small $3\frac{1}{3}$ & 512  & 96 \\ 
        \bottomrule
    \end{tabular}
\label{tab:various_ratios}

\end{table}

\subsection{Alternative Designs}

\begin{figure}[ht] %
\centering
\includegraphics[width=0.5\textwidth]{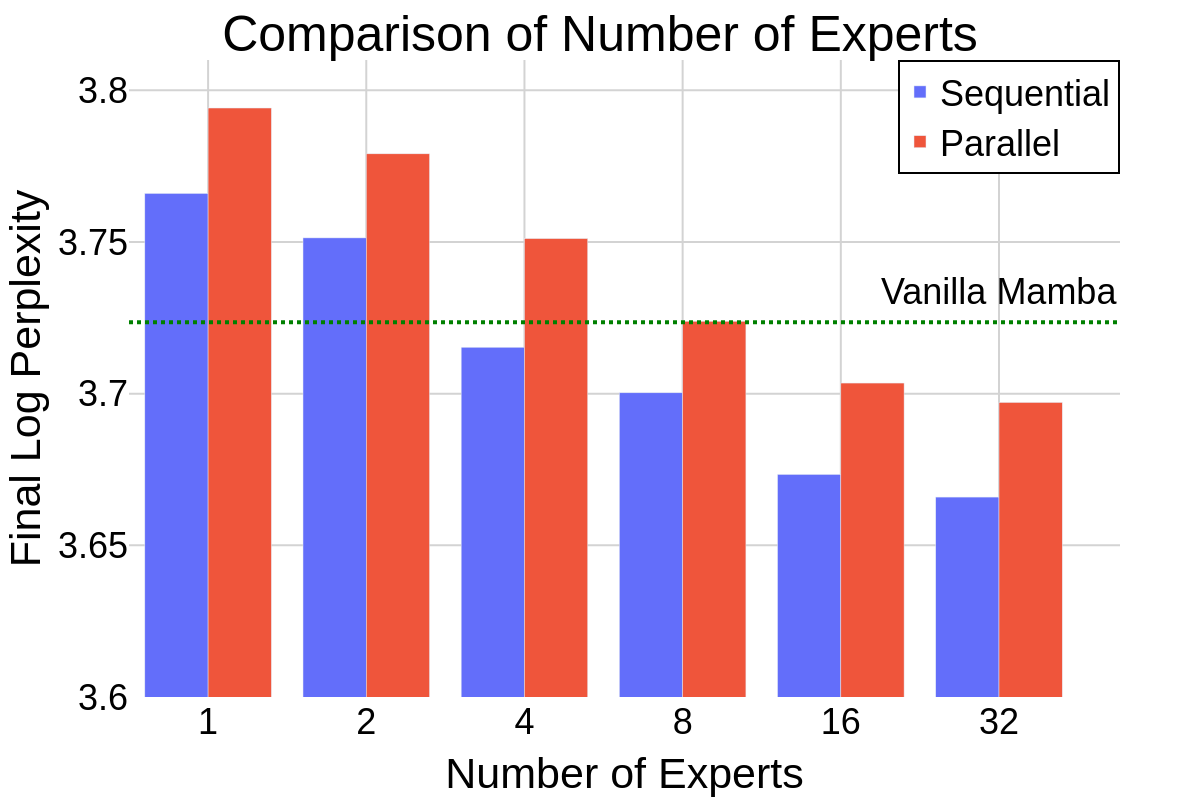}
\caption{Final log perplexity comparison for varying number of experts in sequential and parallel \modelname{}}
\label{fig:parallel_vs_baseline}
\end{figure}

\looseness=-1 \paragraph{Parallel \modelname{}} 
Inspired by \citet{mesh-transformer-jax} and \citet{chowdhery2023palm}, we experiment with an alternative block design in which the MoE \ff layer and the Mamba layer are placed in parallel instead of sequentially (see Figure \ref{fig:other_approaches}). We compare this design to \modelname{} for various numbers of experts; see Figure \ref{fig:parallel_vs_baseline}. \modelname{} outperforms this variant in all tested settings. The parallel \modelname{} matches vanilla Mamba when $N_\text{experts} \geq 8$ while requiring between 2 and 4 times as many experts and total parameters to match the performance of the sequential variant. It may be an attractive alternative at larger scales due to potentially enabling more efficient use of hardware due to different communication \cite{mesh-transformer-jax} or fused input matrix multiplications \cite{chowdhery2023palm}. %

\paragraph{Inner MoE} \label{sec:inner_moe} Pursuing a uniform layer design, we experimented with replacing each of the three linear projections within the Mamba block with an MoE layer; see Figure \ref{fig:other_approaches}. Enumerating all the possible placements results in $2^3-1=7$ possible designs (we discard one combination that would feature no MoE inside the block). We maintain a similar number of total parameters and FLOPs in all models by assuring the total number of expert \ff layers in a block sums up to $24$ regardless of the placement, i.e., the $24$ experts are split evenly between one, two or three MoE's inside the block. Inspired by \citet{fedus2022switch}, we also performed experiments in which only half of the Mamba blocks were modified to include MoE, but the number of experts was increased to $48$ to maintain the total number of parameters. %

Three of the designs (Table \ref{tab:moeinmamba_results}) achieved results marginally better than vanilla Mamba, with none outperforming \modelname. These results suggest the most promising research directions in future work.

\begin{table}[]
    \centering
    \caption{Comparison of different variants of MoE in Mamba - final log perplexity (1B tokens).}
    \begin{tabular}{c|cc}
        \toprule
        \multirow{2}{*}{\thead{\normalsize \textbf{Model Name / }\\ \normalsize \textbf{Modified Projection}}} & \multicolumn{2}{c}{\textbf{MoE in Mamba}} \\
        & \thead{All \\ Layers} & \thead{Every Other \\ Layer} \\
        \midrule 
        Vanilla Mamba & \multicolumn{2}{c}{3.72} \\
        \modelname{} (16 experts) & \multicolumn{2}{c}{\textbf{3.67}} \\
        Conv Projection & 3.79 & 3.71 \\
        Gate Projection & 3.89 & 3.70 \\
        Output Projection & 4.05 & 3.70 \\
        Conv + Gate Projection & 3.95 & 3.72 \\
        Conv + Output Projection & 4.17 & 3.76 \\
        Gate + Output Projection & 4.16 & 3.88 \\
        Conv + Gate + Output Projection & 4.39 & 3.88 \\
        \bottomrule
    \end{tabular}
    \label{tab:moeinmamba_results}
\end{table}

\subsection{Number of Experts} \label{sec:ablations}

\begin{table*}[h!]
    \centering
    \caption{Log perplexity after 1B tokens for various numbers of experts. Note that the parameter counts exclude the embedding and output (unembedding) layers.}
    \begin{tabular}{ccc|cc}
        \toprule
        \textbf{Number of Experts} & \textbf{\# Parameters} & \thead{\textbf{\# Active Parameters} \\ \textbf{per Token}} & \thead{\textbf{Log Perplexity After} \\ \textbf{1B Tokens}} & \thead{ \textbf{Speedup Over} \\ \textbf{Vanilla Mamba}\\ \textbf{(Training Steps)} }  \\
        \midrule
        N/A - Vanilla Mamba & 27M & 27M & 3.72 & 1 \\
        1  & 26M & 26M & 3.75 & $<$1 \\
        4 experts & 64M & 26M & 3.72 & 1.03 \\
        8 experts & 114M & 26M & 3.70 & 1.10 \\
        16 experts & 215M & 26M & 3.67 & 1.21 \\
        32 experts & 416M & 26M & \textbf{3.67} & \textbf{1.23} \\
        \bottomrule
    \end{tabular}

    \label{tab:experts_results}
\end{table*}

Figure \ref{fig:experts_comparison} shows the training runs for different numbers of experts. The results show that our approach scales favorably with the number of experts. \modelname{} outperforms vanilla Mamba, when $N_\text{experts} \geq 4$. We obtain the best result with 32 experts and expect further gains with even more experts.

Interestingly, models with a small number of experts perform worse than vanilla Mamba. This is consistent with \citet{gu2023mamba} reporting that Mamba interleaved with \ff layers (which corresponds to a single-expert MoE layer) is worse than vanilla Mamba.

\subsection{Accuracy and Perplexity} 
\label{sec:acc_perp}
We observed that throughout the training of a variant of one of our smaller models, \moemambaSmall{} with 32 instead of 42 experts as presented in section \ref{sec:main_results}, it maintains a lower perplexity than our strongest baseline (Transformer-MoE). However, at the same time, Transformer-MoE consistently achieves higher accuracy than \modelname. We conjecture that this might be due to the fact that attention-based models are able to copy tokens verbatim, unlike SSM-based models, whose similar abilities might be hindered by the compression of the history into a finite hidden state. We present accuracy and loss (log perplexity) plots alongside further discussion of those results in Appendix \ref{app:acc_perp}.

\section{Future Work and Limitations %
}

\textbf{Scaling}  In this work, we perform experiments on models with the number of active parameters per token smaller than 1B, with total parameters up to 2.4B. Since MoE has enabled Transformers to be scaled to unprecedented sizes \cite{fedus2022switch}, we will be excited to see the impact of scaling on the approaches proposed in our work. Developing scaling laws would be instrumental in this endeavor.

\textbf{Integrating MoE Into the Mamba Layer} Our experiments show that interleaving the Mamba layer with a performant sparse MoE \ff layer results in a promising model. However, in the dense setting, Mamba performs slightly better without the \ff layer. This suggests that integrating sparse computation within the Mamba layer itself could yield even better results while conserving a simple, homogeneous architecture. Our experiments, detailed in section \ref{sec:inner_moe}, warrant some optimism, and we expect this line of research to remain relevant.

\textbf{Exploration of Different Types of MoE in \modelname} 
While we base our design on the commonly used Switch \cite{fedus2022switch}, numerous other MoE architectures have been proposed. Not only may those designs perform better overall, but it is possible that a different type of MoE will be optimal when combined with SSMs. Among possible changes in this regard there are Expert-Choice routers \cite{zhou2022mixtureofexperts}, fully differentiable architectures \cite{puigcerver2023sparse, antoniak2023mixture}, varying number of experts and their granularity, \citep{clark2022unified, krajewski2024scaling}, and other modifications.

\textbf{Distillation} Some works, e.g., \cite{fedus2022switch}, have shown that MoE layers can be distilled back to \ff layers. We expect similar results for \modelname. Interestingly, the findings by \citet{gu2023mamba} indicate that a Mamba module can emulate \ff layers well. This raises the question of whether MoE can be distilled into a vanilla Mamba module and how that would be achieved.

\textbf{Synergies} We leave for future work more in-depth studies of synergies of Mamba and MoE. We suspect that there might be efficiency gains growing with the context length due to better hardware utilization; as for inference, Mamba alleviates computation and memory throughput issues stemming from larger context sizes, while MoE alleviates those same issues stemming from increasing number of parameters and knowledge stored in the model. This synergy may allow for unprecedented scaling of language models both in the number of parameters and length of the input/output. %

\textbf{Mamba and Attention Mechanism} Mamba and Transformers make different trade-offs during data processing. This results in a different set of strengths and weaknesses, e.g., Mamba can process very long inputs but might struggle with tasks requiring detailed knowledge of the past input (e.g., some instances of copying). It would be interesting to explore combining those two architectures to achieve the best of both worlds. %

\textbf{Long Context Utilization} Mamba and other SSMs are praised for their ability to process long context. However, the extent to which they can utilize it effectively and techniques for improving the utilization have not yet been studied in depth. To that end, some methods developed for Transformers \cite{shi2023incontext, tworkowski2023focused, staniszewski2024structured} might be applicable. 

\textbf{Other Modalities} This work explores one direction in which Mamba can be extended. Mamba is a general architecture, and it is not limited to language modeling. We expect that it will be possible to apply \modelname{} to other tasks, like non-textual sequence modeling presented by \citet{gu2023mamba}, and different modalities, such as vision, with initial work presented by \citet{zhu2024vision}.

\section{Conclusions}
In this work, we presented the first integration of Mixture of Experts with Mamba architecture, \modelname. This novel method shares the inference benefits of Mamba while requiring $2.35\times$ fewer training steps to reach the same performance. We showed possible ways of combining those techniques and positively verified performance improvements achieved with their combination. We confirmed with experiments on models up to $2.4$B parameters and training lengths up to $30$B tokens that those improvements over Mamba are robust to model sizes, length of training, and the number of experts.

In addition to the above, we explored and evaluated numerous alternative designs integrating Mixture of Experts within the Mamba block. While none of those variants outperformed \modelname{}, we think that those investigations can help prune ineffective research directions and point to promising ones.

Our work opens a new research direction of combining Mixture of Experts with State Space Models. We believe that this path will enable more efficient scaling to even larger language models.

\newpage
\section*{Acknowledgments}
We would like to express sincere gratitude to Tomasz Odrzygóźdź for the engineering contributions made to our shared repository. We also thank Piotr Sankowski for creating a supportive environment
and providing research direction.

This work was funded by IDEAS NCBR, which also provided significant computational resources and a supportive research environment. The research was supported by PL-Grid infrastructure (grant PLG/2023/016148). We acknowledge snakes and experts as essential to our
work. We also benefited from the Entropy cluster (hosted at the Faculty of Mathematics, Informatics and Mechanics of the University of Warsaw) funded by NVIDIA, Intel, the Polish National Science Center grant 2022/45/N/ST6/02222, and ERC Starting Grant TOTAL. Marek Cygan was partially supported by an NCBiR grant POIR.01.01.01-00-0392/17-00.

\bibliography{main}

\begin{thebibliography}{59}
\providecommand{\natexlab}[1]{#1}
\providecommand{\url}[1]{\texttt{#1}}
\expandafter\ifx\csname urlstyle\endcsname\relax
  \providecommand{\doi}[1]{doi: #1}\else
  \providecommand{\doi}{doi: \begingroup \urlstyle{rm}\Url}\fi

\bibitem[Anthony et~al.(2024)Anthony, Tokpanov, Glorioso, and Millidge]{anthony2024blackmamba}
Anthony, Q., Tokpanov, Y., Glorioso, P., and Millidge, B.
\newblock Blackmamba: Mixture of experts for state-space models, 2024.

\bibitem[Antoniak et~al.(2023)Antoniak, Jaszczur, Krutul, Pióro, Krajewski, Ludziejewski, Odrzygóźdź, and Cygan]{antoniak2023mixture}
Antoniak, S., Jaszczur, S., Krutul, M., Pióro, M., Krajewski, J., Ludziejewski, J., Odrzygóźdź, T., and Cygan, M.
\newblock Mixture of tokens: Efficient llms through cross-example aggregation, 2023.

\bibitem[Artetxe et~al.(2021)Artetxe, Bhosale, Goyal, Mihaylov, Ott, Shleifer, Lin, Du, Iyer, Pasunuru, et~al.]{artetxe2021efficient}
Artetxe, M., Bhosale, S., Goyal, N., Mihaylov, T., Ott, M., Shleifer, S., Lin, X.~V., Du, J., Iyer, S., Pasunuru, R., et~al.
\newblock Efficient large scale language modeling with mixtures of experts.
\newblock \emph{arXiv preprint arXiv:2112.10684}, 2021.

\bibitem[Brown et~al.(2020)Brown, Mann, Ryder, Subbiah, Kaplan, Dhariwal, Neelakantan, Shyam, Sastry, Askell, Agarwal, Herbert{-}Voss, Krueger, Henighan, Child, Ramesh, Ziegler, Wu, Winter, Hesse, Chen, Sigler, Litwin, Gray, Chess, Clark, Berner, McCandlish, Radford, Sutskever, and Amodei]{brown2020language}
Brown, T.~B., Mann, B., Ryder, N., Subbiah, M., Kaplan, J., Dhariwal, P., Neelakantan, A., Shyam, P., Sastry, G., Askell, A., Agarwal, S., Herbert{-}Voss, A., Krueger, G., Henighan, T., Child, R., Ramesh, A., Ziegler, D.~M., Wu, J., Winter, C., Hesse, C., Chen, M., Sigler, E., Litwin, M., Gray, S., Chess, B., Clark, J., Berner, C., McCandlish, S., Radford, A., Sutskever, I., and Amodei, D.
\newblock Language models are few-shot learners.
\newblock \emph{CoRR}, abs/2005.14165, 2020.
\newblock URL \url{https://arxiv.org/abs/2005.14165}.

\bibitem[Chowdhery et~al.(2023)Chowdhery, Narang, Devlin, Bosma, Mishra, Roberts, Barham, Chung, Sutton, Gehrmann, et~al.]{chowdhery2023palm}
Chowdhery, A., Narang, S., Devlin, J., Bosma, M., Mishra, G., Roberts, A., Barham, P., Chung, H.~W., Sutton, C., Gehrmann, S., et~al.
\newblock Palm: Scaling language modeling with pathways.
\newblock \emph{Journal of Machine Learning Research}, 24\penalty0 (240):\penalty0 1--113, 2023.

\bibitem[Clark et~al.(2022)Clark, de~las Casas, Guy, Mensch, Paganini, Hoffmann, Damoc, Hechtman, Cai, Borgeaud, van~den Driessche, Rutherford, Hennigan, Johnson, Millican, Cassirer, Jones, Buchatskaya, Budden, Sifre, Osindero, Vinyals, Rae, Elsen, Kavukcuoglu, and Simonyan]{clark2022unified}
Clark, A., de~las Casas, D., Guy, A., Mensch, A., Paganini, M., Hoffmann, J., Damoc, B., Hechtman, B., Cai, T., Borgeaud, S., van~den Driessche, G., Rutherford, E., Hennigan, T., Johnson, M., Millican, K., Cassirer, A., Jones, C., Buchatskaya, E., Budden, D., Sifre, L., Osindero, S., Vinyals, O., Rae, J., Elsen, E., Kavukcuoglu, K., and Simonyan, K.
\newblock Unified scaling laws for routed language models, 2022.

\bibitem[Devlin et~al.(2019)Devlin, Chang, Lee, and Toutanova]{devlin2019bert}
Devlin, J., Chang, M.-W., Lee, K., and Toutanova, K.
\newblock Bert: Pre-training of deep bidirectional transformers for language understanding, 2019.

\bibitem[Du et~al.(2022)Du, Huang, Dai, Tong, Lepikhin, Xu, Krikun, Zhou, Yu, Firat, et~al.]{du2022glam}
Du, N., Huang, Y., Dai, A.~M., Tong, S., Lepikhin, D., Xu, Y., Krikun, M., Zhou, Y., Yu, A.~W., Firat, O., et~al.
\newblock Glam: Efficient scaling of language models with mixture-of-experts.
\newblock In \emph{International Conference on Machine Learning}, pp.\  5547--5569. PMLR, 2022.

\bibitem[Elhage et~al.(2021)Elhage, Nanda, Olsson, Henighan, Joseph, Mann, Askell, Bai, Chen, Conerly, DasSarma, Drain, Ganguli, Hatfield-Dodds, Hernandez, Jones, Kernion, Lovitt, Ndousse, Amodei, Brown, Clark, Kaplan, McCandlish, and Olah]{elhage2021mathematical}
Elhage, N., Nanda, N., Olsson, C., Henighan, T., Joseph, N., Mann, B., Askell, A., Bai, Y., Chen, A., Conerly, T., DasSarma, N., Drain, D., Ganguli, D., Hatfield-Dodds, Z., Hernandez, D., Jones, A., Kernion, J., Lovitt, L., Ndousse, K., Amodei, D., Brown, T., Clark, J., Kaplan, J., McCandlish, S., and Olah, C.
\newblock A mathematical framework for transformer circuits.
\newblock \emph{Transformer Circuits Thread}, 2021.
\newblock https://transformer-circuits.pub/2021/framework/index.html.

\bibitem[Fedus et~al.(2022)Fedus, Zoph, and Shazeer]{fedus2022switch}
Fedus, W., Zoph, B., and Shazeer, N.
\newblock Switch transformers: Scaling to trillion parameter models with simple and efficient sparsity, 2022.

\bibitem[Fu et~al.(2023)Fu, Dao, Saab, Thomas, Rudra, and Ré]{fu2023hungry}
Fu, D.~Y., Dao, T., Saab, K.~K., Thomas, A.~W., Rudra, A., and Ré, C.
\newblock Hungry hungry hippos: Towards language modeling with state space models, 2023.

\bibitem[Gu \& Dao(2023)Gu and Dao]{gu2023mamba}
Gu, A. and Dao, T.
\newblock Mamba: Linear-time sequence modeling with selective state spaces, 2023.

\bibitem[Gu et~al.(2021)Gu, Johnson, Goel, Saab, Dao, Rudra, and Ré]{gu2021combining}
Gu, A., Johnson, I., Goel, K., Saab, K., Dao, T., Rudra, A., and Ré, C.
\newblock Combining recurrent, convolutional, and continuous-time models with linear state-space layers, 2021.

\bibitem[Gu et~al.(2022{\natexlab{a}})Gu, Goel, Gupta, and R{\'e}]{gu2022parameterization}
Gu, A., Goel, K., Gupta, A., and R{\'e}, C.
\newblock On the parameterization and initialization of diagonal state space models.
\newblock \emph{Advances in Neural Information Processing Systems}, 35:\penalty0 35971--35983, 2022{\natexlab{a}}.

\bibitem[Gu et~al.(2022{\natexlab{b}})Gu, Goel, and Ré]{gu2022efficiently}
Gu, A., Goel, K., and Ré, C.
\newblock Efficiently modeling long sequences with structured state spaces, 2022{\natexlab{b}}.

\bibitem[Gupta et~al.(2022)Gupta, Gu, and Berant]{gupta2022diagonal}
Gupta, A., Gu, A., and Berant, J.
\newblock Diagonal state spaces are as effective as structured state spaces.
\newblock \emph{Advances in Neural Information Processing Systems}, 35:\penalty0 22982--22994, 2022.

\bibitem[Jacobs et~al.(1991)Jacobs, Jordan, Nowlan, and Hinton]{moe1991}
Jacobs, R.~A., Jordan, M.~I., Nowlan, S.~J., and Hinton, G.~E.
\newblock Adaptive mixtures of local experts.
\newblock \emph{Neural Computation}, 3\penalty0 (1):\penalty0 79--87, 1991.
\newblock \doi{10.1162/neco.1991.3.1.79}.

\bibitem[Jiang et~al.(2024)Jiang, Sablayrolles, Roux, Mensch, Savary, Bamford, Chaplot, de~las Casas, Hanna, Bressand, Lengyel, Bour, Lample, Lavaud, Saulnier, Lachaux, Stock, Subramanian, Yang, Antoniak, Scao, Gervet, Lavril, Wang, Lacroix, and Sayed]{jiang2024mixtral}
Jiang, A.~Q., Sablayrolles, A., Roux, A., Mensch, A., Savary, B., Bamford, C., Chaplot, D.~S., de~las Casas, D., Hanna, E.~B., Bressand, F., Lengyel, G., Bour, G., Lample, G., Lavaud, L.~R., Saulnier, L., Lachaux, M.-A., Stock, P., Subramanian, S., Yang, S., Antoniak, S., Scao, T.~L., Gervet, T., Lavril, T., Wang, T., Lacroix, T., and Sayed, W.~E.
\newblock Mixtral of experts, 2024.

\bibitem[Jordan \& Jacobs(1993)Jordan and Jacobs]{716791}
Jordan, M. and Jacobs, R.
\newblock Hierarchical mixtures of experts and the em algorithm.
\newblock In \emph{Proceedings of 1993 International Conference on Neural Networks (IJCNN-93-Nagoya, Japan)}, volume~2, pp.\  1339--1344 vol.2, 1993.
\newblock \doi{10.1109/IJCNN.1993.716791}.

\bibitem[Kaplan et~al.(2020)Kaplan, McCandlish, Henighan, Brown, Chess, Child, Gray, Radford, Wu, and Amodei]{kaplan2020scaling}
Kaplan, J., McCandlish, S., Henighan, T., Brown, T.~B., Chess, B., Child, R., Gray, S., Radford, A., Wu, J., and Amodei, D.
\newblock Scaling laws for neural language models, 2020.

\bibitem[Krajewski et~al.(2024)Krajewski, Ludziejewski, Adamczewski, Pióro, Krutul, Antoniak, Ciebiera, Król, Odrzygóźdź, Sankowski, Cygan, and Jaszczur]{krajewski2024scaling}
Krajewski, J., Ludziejewski, J., Adamczewski, K., Pióro, M., Krutul, M., Antoniak, S., Ciebiera, K., Król, K., Odrzygóźdź, T., Sankowski, P., Cygan, M., and Jaszczur, S.
\newblock Scaling laws for fine-grained mixture of experts, 2024.

\bibitem[Lan et~al.(2020)Lan, Chen, Goodman, Gimpel, Sharma, and Soricut]{lan2020albert}
Lan, Z., Chen, M., Goodman, S., Gimpel, K., Sharma, P., and Soricut, R.
\newblock Albert: A lite bert for self-supervised learning of language representations, 2020.

\bibitem[Lepikhin et~al.(2020)Lepikhin, Lee, Xu, Chen, Firat, Huang, Krikun, Shazeer, and Chen]{lepikhin2020gshard}
Lepikhin, D., Lee, H., Xu, Y., Chen, D., Firat, O., Huang, Y., Krikun, M., Shazeer, N., and Chen, Z.
\newblock Gshard: Scaling giant models with conditional computation and automatic sharding, 2020.

\bibitem[Lewkowycz et~al.(2022)Lewkowycz, Andreassen, Dohan, Dyer, Michalewski, Ramasesh, Slone, Anil, Schlag, Gutman-Solo, Wu, Neyshabur, Gur-Ari, and Misra]{lewkowycz2022solving}
Lewkowycz, A., Andreassen, A.~J., Dohan, D., Dyer, E., Michalewski, H., Ramasesh, V.~V., Slone, A., Anil, C., Schlag, I., Gutman-Solo, T., Wu, Y., Neyshabur, B., Gur-Ari, G., and Misra, V.
\newblock Solving quantitative reasoning problems with language models.
\newblock In Oh, A.~H., Agarwal, A., Belgrave, D., and Cho, K. (eds.), \emph{Advances in Neural Information Processing Systems}, 2022.
\newblock URL \url{https://openreview.net/forum?id=IFXTZERXdM7}.

\bibitem[Li et~al.(2022)Li, Cai, Zhang, Chen, and Dey]{li2022makes}
Li, Y., Cai, T., Zhang, Y., Chen, D., and Dey, D.
\newblock What makes convolutional models great on long sequence modeling?
\newblock \emph{arXiv preprint arXiv:2210.09298}, 2022.

\bibitem[Loshchilov \& Hutter(2019)Loshchilov and Hutter]{loshchilov2019decoupled}
Loshchilov, I. and Hutter, F.
\newblock Decoupled weight decay regularization, 2019.

\bibitem[Ma et~al.(2022)Ma, Zhou, Kong, He, Gui, Neubig, May, and Zettlemoyer]{ma2022mega}
Ma, X., Zhou, C., Kong, X., He, J., Gui, L., Neubig, G., May, J., and Zettlemoyer, L.
\newblock Mega: moving average equipped gated attention.
\newblock \emph{arXiv preprint arXiv:2209.10655}, 2022.

\bibitem[Olsson et~al.(2022)Olsson, Elhage, Nanda, Joseph, DasSarma, Henighan, Mann, Askell, Bai, Chen, Conerly, Drain, Ganguli, Hatfield-Dodds, Hernandez, Johnston, Jones, Kernion, Lovitt, Ndousse, Amodei, Brown, Clark, Kaplan, McCandlish, and Olah]{olsson2022context}
Olsson, C., Elhage, N., Nanda, N., Joseph, N., DasSarma, N., Henighan, T., Mann, B., Askell, A., Bai, Y., Chen, A., Conerly, T., Drain, D., Ganguli, D., Hatfield-Dodds, Z., Hernandez, D., Johnston, S., Jones, A., Kernion, J., Lovitt, L., Ndousse, K., Amodei, D., Brown, T., Clark, J., Kaplan, J., McCandlish, S., and Olah, C.
\newblock In-context learning and induction heads.
\newblock \emph{Transformer Circuits Thread}, 2022.
\newblock https://transformer-circuits.pub/2022/in-context-learning-and-induction-heads/index.html.

\bibitem[OpenAI(2023)]{openai2023gpt4}
OpenAI.
\newblock Gpt-4 technical report, 2023.

\bibitem[Orvieto et~al.(2023)Orvieto, Smith, Gu, Fernando, Gulcehre, Pascanu, and De]{orvieto2023resurrecting}
Orvieto, A., Smith, S.~L., Gu, A., Fernando, A., Gulcehre, C., Pascanu, R., and De, S.
\newblock Resurrecting recurrent neural networks for long sequences.
\newblock \emph{arXiv preprint arXiv:2303.06349}, 2023.

\bibitem[Paszke et~al.(2019)Paszke, Gross, Massa, Lerer, Bradbury, Chanan, Killeen, Lin, Gimelshein, Antiga, Desmaison, Köpf, Yang, DeVito, Raison, Tejani, Chilamkurthy, Steiner, Fang, Bai, and Chintala]{paszke2019pytorch}
Paszke, A., Gross, S., Massa, F., Lerer, A., Bradbury, J., Chanan, G., Killeen, T., Lin, Z., Gimelshein, N., Antiga, L., Desmaison, A., Köpf, A., Yang, E., DeVito, Z., Raison, M., Tejani, A., Chilamkurthy, S., Steiner, B., Fang, L., Bai, J., and Chintala, S.
\newblock Pytorch: An imperative style, high-performance deep learning library, 2019.

\bibitem[Peng et~al.(2023)Peng, Alcaide, Anthony, Albalak, Arcadinho, Biderman, Cao, Cheng, Chung, Grella, GV, He, Hou, Lin, Kazienko, Kocon, Kong, Koptyra, Lau, Mantri, Mom, Saito, Song, Tang, Wang, Wind, Wozniak, Zhang, Zhang, Zhao, Zhou, Zhou, Zhu, and Zhu]{peng2023rwkv}
Peng, B., Alcaide, E., Anthony, Q., Albalak, A., Arcadinho, S., Biderman, S., Cao, H., Cheng, X., Chung, M., Grella, M., GV, K.~K., He, X., Hou, H., Lin, J., Kazienko, P., Kocon, J., Kong, J., Koptyra, B., Lau, H., Mantri, K. S.~I., Mom, F., Saito, A., Song, G., Tang, X., Wang, B., Wind, J.~S., Wozniak, S., Zhang, R., Zhang, Z., Zhao, Q., Zhou, P., Zhou, Q., Zhu, J., and Zhu, R.-J.
\newblock Rwkv: Reinventing rnns for the transformer era, 2023.

\bibitem[Poli et~al.(2023)Poli, Massaroli, Nguyen, Fu, Dao, Baccus, Bengio, Ermon, and Ré]{poli2023hyena}
Poli, M., Massaroli, S., Nguyen, E., Fu, D.~Y., Dao, T., Baccus, S., Bengio, Y., Ermon, S., and Ré, C.
\newblock Hyena hierarchy: Towards larger convolutional language models, 2023.

\bibitem[Puigcerver et~al.(2023)Puigcerver, Riquelme, Mustafa, and Houlsby]{puigcerver2023sparse}
Puigcerver, J., Riquelme, C., Mustafa, B., and Houlsby, N.
\newblock From sparse to soft mixtures of experts, 2023.

\bibitem[Radford et~al.(2019)Radford, Wu, Child, Luan, Amodei, Sutskever, et~al.]{radford2019language}
Radford, A., Wu, J., Child, R., Luan, D., Amodei, D., Sutskever, I., et~al.
\newblock Language models are unsupervised multitask learners.
\newblock \emph{OpenAI blog}, 1\penalty0 (8):\penalty0 9, 2019.

\bibitem[Raffel et~al.(2020)Raffel, Shazeer, Roberts, Lee, Narang, Matena, Zhou, Li, and Liu]{raffel2020exploring}
Raffel, C., Shazeer, N., Roberts, A., Lee, K., Narang, S., Matena, M., Zhou, Y., Li, W., and Liu, P.~J.
\newblock Exploring the limits of transfer learning with a unified text-to-text transformer.
\newblock \emph{The Journal of Machine Learning Research}, 21\penalty0 (1):\penalty0 5485--5551, 2020.

\bibitem[Sanseviero et~al.(2023)Sanseviero, Tunstall, Schmid, Mangrulkar, Belkada, and Cuenca]{sanseviero2023moe}
Sanseviero, O., Tunstall, L., Schmid, P., Mangrulkar, S., Belkada, Y., and Cuenca, P.
\newblock Mixture of experts explained, 2023.
\newblock URL \url{https://huggingface.co/blog/moe}.

\bibitem[Shazeer et~al.(2017)Shazeer, Mirhoseini, Maziarz, Davis, Le, Hinton, and Dean]{shazeer2017outrageously}
Shazeer, N., Mirhoseini, A., Maziarz, K., Davis, A., Le, Q., Hinton, G., and Dean, J.
\newblock Outrageously large neural networks: The sparsely-gated mixture-of-experts layer, 2017.

\bibitem[Shi et~al.(2023)Shi, Min, Lomeli, Zhou, Li, James, Lin, Smith, Zettlemoyer, Yih, and Lewis]{shi2023incontext}
Shi, W., Min, S., Lomeli, M., Zhou, C., Li, M., James, R., Lin, X.~V., Smith, N.~A., Zettlemoyer, L., Yih, S., and Lewis, M.
\newblock In-context pretraining: Language modeling beyond document boundaries, 2023.

\bibitem[Smith et~al.(2023)Smith, Warrington, and Linderman]{smith2023simplified}
Smith, J. T.~H., Warrington, A., and Linderman, S.~W.
\newblock Simplified state space layers for sequence modeling, 2023.

\bibitem[Staniszewski et~al.(2024)Staniszewski, Tworkowski, Jaszczur, Michalewski, Łukasz Kuciński, and Miłoś]{staniszewski2024structured}
Staniszewski, K., Tworkowski, S., Jaszczur, S., Michalewski, H., Łukasz Kuciński, and Miłoś, P.
\newblock Structured packing in llm training improves long context utilization, 2024.

\bibitem[Su et~al.(2023)Su, Lu, Pan, Murtadha, Wen, and Liu]{su2023roformer}
Su, J., Lu, Y., Pan, S., Murtadha, A., Wen, B., and Liu, Y.
\newblock Roformer: Enhanced transformer with rotary position embedding, 2023.

\bibitem[Sun et~al.(2023)Sun, Dong, Huang, Ma, Xia, Xue, Wang, and Wei]{sun2023retentive}
Sun, Y., Dong, L., Huang, S., Ma, S., Xia, Y., Xue, J., Wang, J., and Wei, F.
\newblock Retentive network: A successor to transformer for large language models, 2023.

\bibitem[Sutton(2019)]{sutton2019bitter}
Sutton, R.
\newblock The bitter lesson.
\newblock \emph{Incomplete Ideas (blog)}, 13\penalty0 (1), 2019.

\bibitem[Tay et~al.(2020)Tay, Dehghani, Abnar, Shen, Bahri, Pham, Rao, Yang, Ruder, and Metzler]{tay2020long}
Tay, Y., Dehghani, M., Abnar, S., Shen, Y., Bahri, D., Pham, P., Rao, J., Yang, L., Ruder, S., and Metzler, D.
\newblock Long range arena: A benchmark for efficient transformers.
\newblock \emph{arXiv preprint arXiv:2011.04006}, 2020.

\bibitem[Team(2023)]{geminiteam2023gemini}
Team, G.
\newblock Gemini: A family of highly capable multimodal models, 2023.

\bibitem[TogetherComputer(2023)]{together2023redpajama}
TogetherComputer.
\newblock Redpajama: An open source recipe to reproduce llama training dataset, April 2023.
\newblock URL \url{https://github.com/togethercomputer/RedPajama-Data}.

\bibitem[Touvron et~al.(2023)Touvron, Martin, Stone, Albert, Almahairi, Babaei, Bashlykov, Batra, Bhargava, Bhosale, Bikel, Blecher, Ferrer, Chen, Cucurull, Esiobu, Fernandes, Fu, Fu, Fuller, Gao, Goswami, Goyal, Hartshorn, Hosseini, Hou, Inan, Kardas, Kerkez, Khabsa, Kloumann, Korenev, Koura, Lachaux, Lavril, Lee, Liskovich, Lu, Mao, Martinet, Mihaylov, Mishra, Molybog, Nie, Poulton, Reizenstein, Rungta, Saladi, Schelten, Silva, Smith, Subramanian, Tan, Tang, Taylor, Williams, Kuan, Xu, Yan, Zarov, Zhang, Fan, Kambadur, Narang, Rodriguez, Stojnic, Edunov, and Scialom]{touvron2023llama}
Touvron, H., Martin, L., Stone, K., Albert, P., Almahairi, A., Babaei, Y., Bashlykov, N., Batra, S., Bhargava, P., Bhosale, S., Bikel, D., Blecher, L., Ferrer, C.~C., Chen, M., Cucurull, G., Esiobu, D., Fernandes, J., Fu, J., Fu, W., Fuller, B., Gao, C., Goswami, V., Goyal, N., Hartshorn, A., Hosseini, S., Hou, R., Inan, H., Kardas, M., Kerkez, V., Khabsa, M., Kloumann, I., Korenev, A., Koura, P.~S., Lachaux, M.-A., Lavril, T., Lee, J., Liskovich, D., Lu, Y., Mao, Y., Martinet, X., Mihaylov, T., Mishra, P., Molybog, I., Nie, Y., Poulton, A., Reizenstein, J., Rungta, R., Saladi, K., Schelten, A., Silva, R., Smith, E.~M., Subramanian, R., Tan, X.~E., Tang, B., Taylor, R., Williams, A., Kuan, J.~X., Xu, P., Yan, Z., Zarov, I., Zhang, Y., Fan, A., Kambadur, M., Narang, S., Rodriguez, A., Stojnic, R., Edunov, S., and Scialom, T.
\newblock Llama 2: Open foundation and fine-tuned chat models, 2023.

\bibitem[Turc et~al.(2019)Turc, Chang, Lee, and Toutanova]{turc2019wellread}
Turc, I., Chang, M.-W., Lee, K., and Toutanova, K.
\newblock Well-read students learn better: On the importance of pre-training compact models, 2019.

\bibitem[Tworkowski et~al.(2023)Tworkowski, Staniszewski, Pacek, Wu, Michalewski, and Miłoś]{tworkowski2023focused}
Tworkowski, S., Staniszewski, K., Pacek, M., Wu, Y., Michalewski, H., and Miłoś, P.
\newblock Focused transformer: Contrastive training for context scaling, 2023.

\bibitem[Vaswani et~al.(2017)Vaswani, Shazeer, Parmar, Uszkoreit, Jones, Gomez, Kaiser, and Polosukhin]{vaswani2017attention}
Vaswani, A., Shazeer, N., Parmar, N., Uszkoreit, J., Jones, L., Gomez, A.~N., Kaiser, L., and Polosukhin, I.
\newblock Attention is all you need.
\newblock \emph{CoRR}, abs/1706.03762, 2017.
\newblock URL \url{http://arxiv.org/abs/1706.03762}.

\bibitem[Wang(2021)]{mesh-transformer-jax}
Wang, B.
\newblock {Mesh-Transformer-JAX: Model-Parallel Implementation of Transformer Language Model with JAX}.
\newblock \url{https://github.com/kingoflolz/mesh-transformer-jax}, May 2021.

\bibitem[Xue et~al.(2023)Xue, Zheng, Fu, Ni, Zheng, Zhou, and You]{openmoe2023}
Xue, F., Zheng, Z., Fu, Y., Ni, J., Zheng, Z., Zhou, W., and You, Y.
\newblock Openmoe: Open mixture-of-experts language models.
\newblock \url{https://github.com/XueFuzhao/OpenMoE}, 2023.

\bibitem[Zhai et~al.(2021)Zhai, Talbott, Srivastava, Huang, Goh, Zhang, and Susskind]{zhai2021attention}
Zhai, S., Talbott, W., Srivastava, N., Huang, C., Goh, H., Zhang, R., and Susskind, J.
\newblock An attention free transformer, 2021.

\bibitem[Zhao et~al.(2023{\natexlab{a}})Zhao, Zhou, Li, Tang, Wang, Hou, Min, Zhang, Zhang, Dong, Du, Yang, Chen, Chen, Jiang, Ren, Li, Tang, Liu, Liu, Nie, and Wen]{zhao2023survey}
Zhao, W.~X., Zhou, K., Li, J., Tang, T., Wang, X., Hou, Y., Min, Y., Zhang, B., Zhang, J., Dong, Z., Du, Y., Yang, C., Chen, Y., Chen, Z., Jiang, J., Ren, R., Li, Y., Tang, X., Liu, Z., Liu, P., Nie, J.-Y., and Wen, J.-R.
\newblock A survey of large language models, 2023{\natexlab{a}}.

\bibitem[Zhao et~al.(2023{\natexlab{b}})Zhao, Gu, Varma, Luo, Huang, Xu, Wright, Shojanazeri, Ott, Shleifer, Desmaison, Balioglu, Damania, Nguyen, Chauhan, Hao, Mathews, and Li]{zhao2023pytorch}
Zhao, Y., Gu, A., Varma, R., Luo, L., Huang, C.-C., Xu, M., Wright, L., Shojanazeri, H., Ott, M., Shleifer, S., Desmaison, A., Balioglu, C., Damania, P., Nguyen, B., Chauhan, G., Hao, Y., Mathews, A., and Li, S.
\newblock Pytorch fsdp: Experiences on scaling fully sharded data parallel, 2023{\natexlab{b}}.

\bibitem[Zhou et~al.(2022)Zhou, Lei, Liu, Du, Huang, Zhao, Dai, Chen, Le, and Laudon]{zhou2022mixtureofexperts}
Zhou, Y., Lei, T., Liu, H., Du, N., Huang, Y., Zhao, V., Dai, A., Chen, Z., Le, Q., and Laudon, J.
\newblock Mixture-of-experts with expert choice routing, 2022.

\bibitem[Zhu et~al.(2024)Zhu, Liao, Zhang, Wang, Liu, and Wang]{zhu2024vision}
Zhu, L., Liao, B., Zhang, Q., Wang, X., Liu, W., and Wang, X.
\newblock Vision mamba: Efficient visual representation learning with bidirectional state space model, 2024.

\bibitem[Zoph et~al.(2022)Zoph, Bello, Kumar, Du, Huang, Dean, Shazeer, and Fedus]{zoph2022stmoe}
Zoph, B., Bello, I., Kumar, S., Du, N., Huang, Y., Dean, J., Shazeer, N., and Fedus, W.
\newblock St-moe: Designing stable and transferable sparse expert models, 2022.

\end{thebibliography}
\bibliographystyle{icml2024}

\newpage

\appendix
\onecolumn

\section{Hyperparameters and Training Setup} \label{app:hyperparams}
Basic model hyperameters ($\dmodel$, $d_{\text{ff}}$, the number of attention heads, the number of layers) used in this work were inspired by BERT \cite{devlin2019bert,turc2019wellread}, with the \smalll{} models being equivalent to $\text{BERT}_\textsc{Medium}$ and \largee{} models copying $\text{BERT}_\textsc{Base}$ configuration while increasing the number of blocks from $12$ to $16$. The learning rate schedule, as well as weight decay and gradient clipping values were set per community's standard practices. We used the AdamW optimizer \cite{loshchilov2019decoupled}.
We tune the maximum learning rate value for each of the \smalll{} models separately and divide it by 2 when training \largee{} counterparts. 
We train the models using PyTorch \cite{paszke2019pytorch} and utilize FSDP \cite{zhao2023pytorch} for facilitating multi-GPU setup.

\begin{table}[h]
    
    \caption{Hyperparameters (\smalll{} Models). In Transformer models we use Rotary Position Embedding \cite{su2023roformer}.}
    \label{tab:arch_details_small}
    \begin{center}
    \begin{tabular}{c c|c c c c}
        \toprule
        \multicolumn{2}{c|}{\textbf{Hyperparameter}} & \textbf{\transformerSmall{}} & \textbf{\mambaSmall{}} & \textbf{\moeSmall{}} & \textbf{\moemambaSmall{}}  \\
        \midrule
        \multirow{4}{*}{Model} & Total Blocks & 8 & 16 & 8 & 8  \\
                               & $\dmodel$ & 512 & 512 & 512 & 512 \\ 
                               & \# Parameters & 25M & 27M & 545M & 542M \\
                               & \thead{\# Active Parameters \\ per Token} & 25M & 27M & 25M & 26M \\
        \midrule
        \FF & $d_{\text{ff}}$ & 2048 & - & - & - \\
        \midrule
        \multirow{2}{*}{Mixture of Experts} & $d_{\text{expert}}$  & - & - & 2048 & 1536 \\
                                            & $\Nexperts$ & - & - & 32 & 42 \\
        \midrule
        \multicolumn{2}{c|}{Position Embedding} & RoPE & - & RoPE & - \\
        \midrule

        \multirow{1}{*}{Attention} & $N_{\text{heads}}$  & 8 & - & 8 & - \\
        \midrule

        \multirow{9}{*}{Training} & Training Steps  & 150K & 150K & 150K & 150K  \\
                                  & Context Length & 1024 & 1024 & 1024 & 1024 \\
                                  & Batch Size & 64 & 64 & 64 & 64 \\
                                  & Max Learning Rate & 5e-4 & 1e-3 & 5e-4 & 5e-4 \\
                                  & LR Warmup & 1\% & 1\% & 1\% & 1\% \\
                                  & LR Schedule & Cosine & Cosine & Cosine & Cosine \\
                                  & Final LR Ratio & 0.1 & 0.1 & 0.1 & 0.1 \\ 
                                  & Weight Decay & 0.1 & 0.1 & 0.1 & 0.1 \\
                                  & Gradient Clipping & 0.5 & 0.5 & 0.5 & 0.5 \\
        \bottomrule
    \end{tabular}
    \end{center}
\end{table}

\begin{table}[h!]
    
    \label{tab:arch_details_large}
    \begin{center}
    \caption{Hyperparameters (\largee{} Models). In \moeLarge{} we use Rotary Position Embedding \cite{su2023roformer}.}
    
    \begin{tabular}{c c|c c c}
        
        \toprule
        \multicolumn{2}{c|}{\textbf{Hyperparameter}} &  \textbf{\mambaLarge{}} & \textbf{\moeLarge{}} & \textbf{\moemambaLarge{}}  \\
        \midrule
        \multirow{4}{*}{Model} & Total Blocks &  32 & 16 & 16  \\
                               & $\dmodel$ & 768 & 768 & 768 \\ 
                               & \# Parameters & 121M & 2454M & 2439M \\
                               & \thead{\# Active Parameters \\ per Token} & 121M & 114M & 117M \\
        \midrule
        \multirow{2}{*}{Mixture of Experts} & $d_{\text{expert}}$  & - & 3072 & 2304 \\
                                            & $\Nexperts$ & - & 32 & 42 \\
        \midrule
        \multicolumn{2}{c|}{Position Embedding} & - & RoPE & - \\
        \midrule
        \multirow{1}{*}{Attention} & $N_{\text{heads}}$  & - & 12 & - \\
        \midrule

        \multirow{9}{*}{Training} & Training Steps  &  30K & 30K & 30K  \\
                                  & Context Length &  1024 & 1024 & 1024 \\
                                  & Batch Size & 1024 & 1024 & 1024 \\
                                  & Max Learning Rate &  1e-3 & 2.5e-4 & 5e-4 \\
                                  & LR Warmup & 1\% & 1\% & 1\% \\
                                  & LR Schedule &  Cosine & Cosine & Cosine \\
                                  & Final LR Ratio & 0.1 & 0.1 & 0.1 \\ 
                                  & Weight Decay &  0.1 & 0.1 & 0.1 \\
                                  & Gradient Clipping &  0.5 & 0.5 & 0.5 \\
        \bottomrule
    \end{tabular}
    \end{center}
\end{table}

\section{Active Parameters vs FLOPs} \label{app:active_params}

In this work, we report the number of active parameters (excluding embedding and unembedding layers) and not the number of floating-point operations (FLOPs), following \cite{zhou2022mixtureofexperts}. Both numbers will be roughly proportional \cite{kaplan2020scaling}, but the number of FLOPs is both harder to calculate and less relevant for hardware-aware architecture like Mamba with its optimizations, especially during inference.

\begin{table}[]
    \caption{Comparison of sequential and parallel \modelname{} - final log perplexity (1B tokens).}
    \centering
    \begin{tabular}{c|cc}
        \toprule
        \multirow{2}{*}{\textbf{\# of Experts}} & \multicolumn{2}{c}{\textbf{\modelname{}}} \\
        & Sequential & Parallel \\
        \midrule 
        1 & 3.76 & 3.79 \\
        2 & 3.74 & 3.77 \\
        4 & 3.71 & 3.74 \\
        8 & 3.69 & 3.72 \\
        16 & 3.67 & 3.70 \\
        32 & \textbf{3.66} & 3.69 \\
        \bottomrule
    \end{tabular}
    \label{tab:parallel_results}
\end{table}

\section{Accuracy and Perplexity} \label{app:acc_perp}

\begin{figure}[H]
    \centering
    \includegraphics[width=0.495\textwidth]{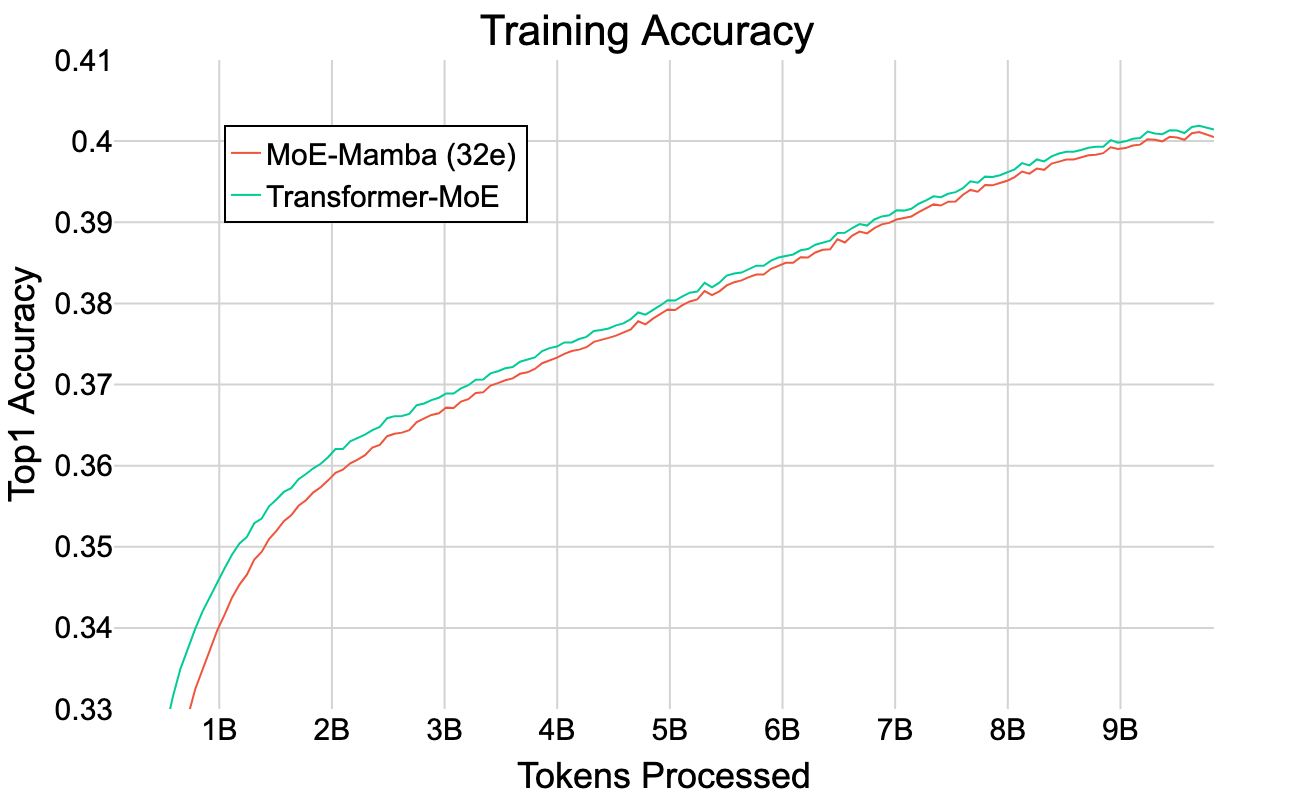}
    \includegraphics[width=0.495\textwidth]{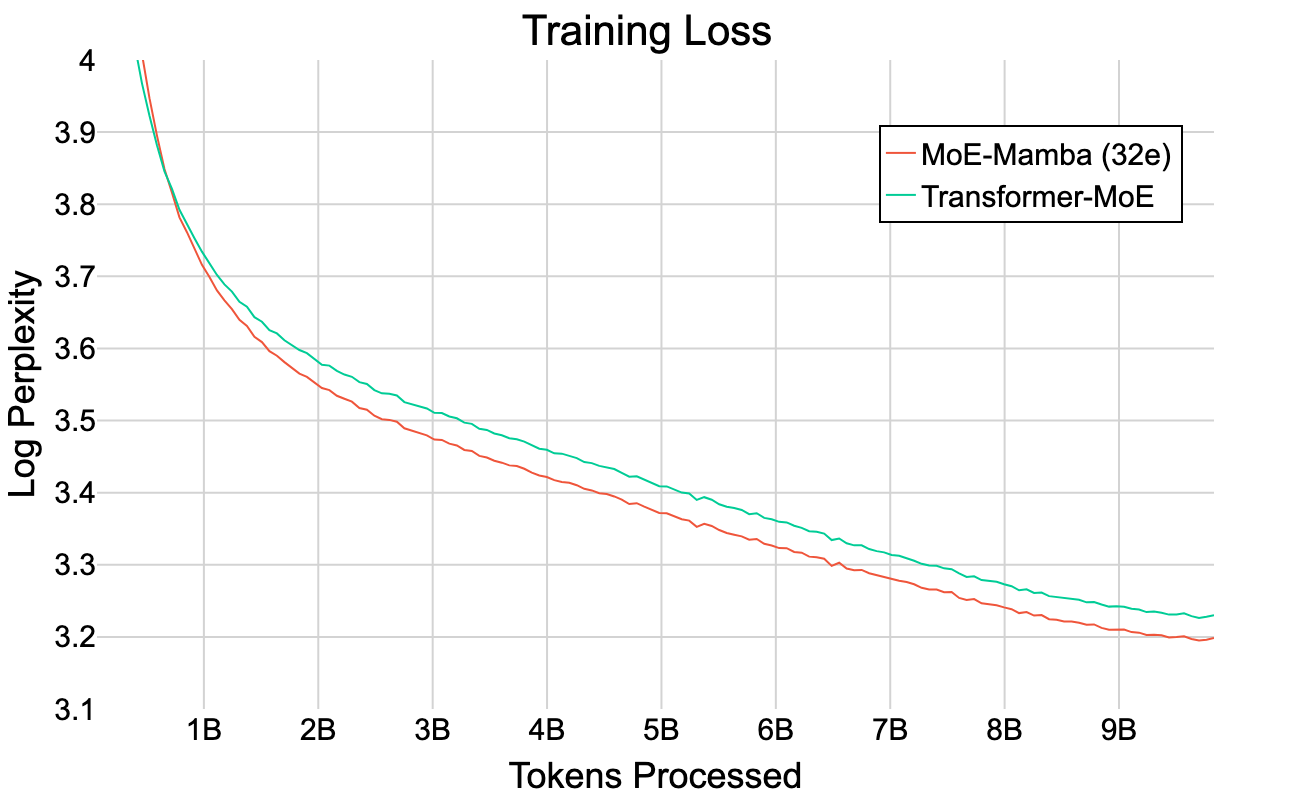}
    \caption{Discrepancy between accuracy and log perplexity: \moemambaSmall{} with 32 experts and \transformerSmall. Note that MoE-Mamba with 32 experts has fewer total parameters than the Transformer.}
\end{figure}

As mentioned in section \ref{sec:acc_perp}, we have observed a curious case of metric inconsistency between two models that achieved similar performance but were based on different architectures.
We hypothesize that this discrepancy hints at a potential failure mode of Mamba and other SSMs. Due to the compression of the history into a finite hidden state, their ability for verbatim token-copying is limited. The related ability to predict the token $[B]$ given a prefix $...[A][B]...[A]$ (where $[A], [B]$ can be any tokens) has been mechanistically studied by \citet{elhage2021mathematical} and has been conjectured to be responsible for Transformer's remarkable in-context learning capabilities \cite{olsson2022context}. 

\citet{peng2023rwkv} mention that their attention-free model, RWKV, may have limited performance on tasks that require recalling precise information over long contexts due to a fixed-sized hidden state, a property that Mamba and other SSMs share. However, since the perplexity of Mamba can match the perplexity of a similarly-sized Transformer, we can suspect that Mamba compensates for that failure mode in other ways and might show a relative advantage on other tasks when compared to Transformer. In particular, it might outperform Transformers in 0-shot tasks in contrast to tasks allowing few-shot demonstrations or requiring in-context learning.

\section{Relation between Speedup and Training Time} \label{app:speedup}
\begin{figure}[ht]
\centering
\includegraphics[width=0.8\textwidth]{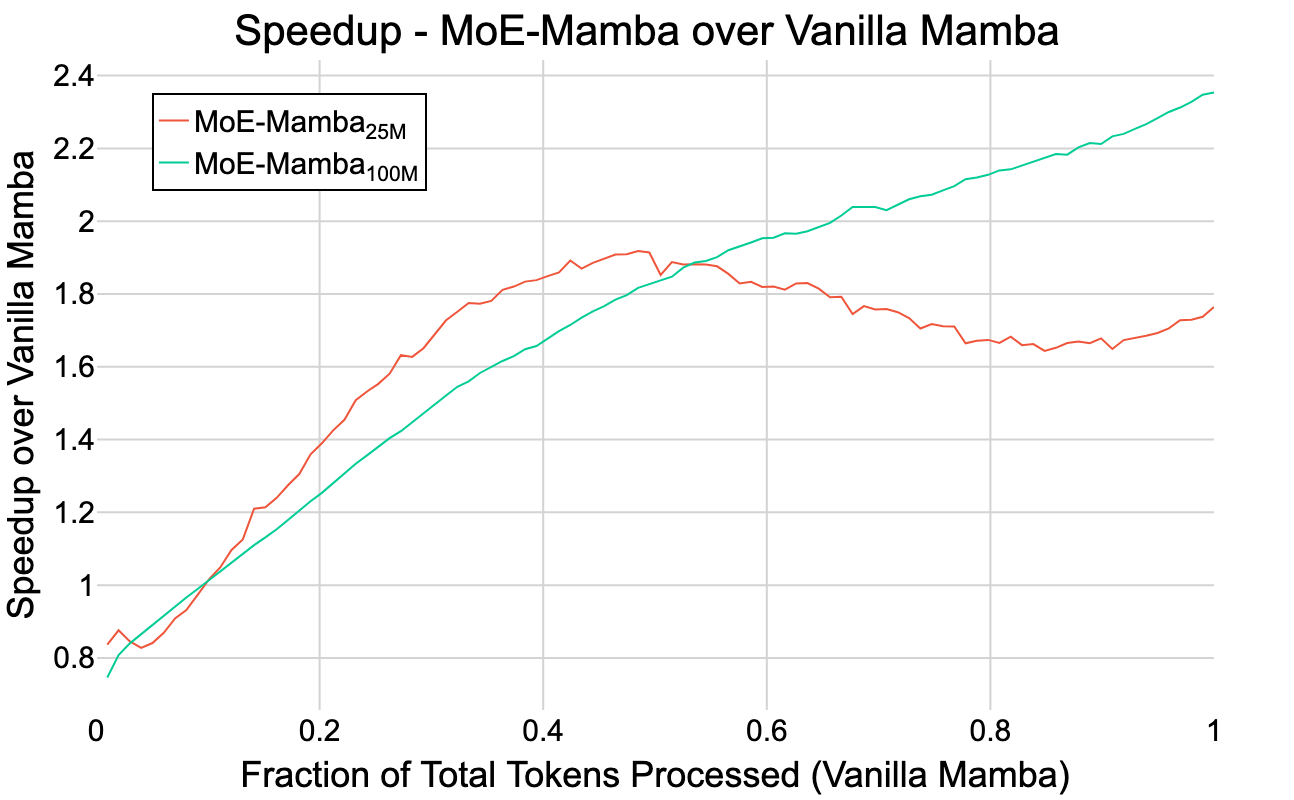}
\caption{Speedup of different sizes of \modelname{} compared to their vanilla Mamba counterparts as training progresses.} %
\label{fig:speedup}
\end{figure}

In our experiments, we notice that generally, as the training continues, the speedup of \modelname{} compared to vanilla Mamba increases (see Fig. \ref{fig:speedup}).
That is, the ratio
$$
\text{speedup}(l) = \frac{\text{ \# processed tokens vanilla Mamba took to reach loss } l}{\text{\# processed tokens \modelname{} took to reach loss } l}
$$
increases as $l$ decreases. 
Speedup in \smalll{} models oscillates between 1.6 and 1.9, while the speedup in \largee{} models rises steadily.

\section{Counting Model Parameters} \label{app:params}
For all models and their variants, we report the number of trainable, non-embedding parameters, i.e., we exclude the parameters in the input (embedding) and output (unembedding) layers. This convention is proposed by \citet{kaplan2020scaling}, who note that using just non-embedding parameters gives their scaling laws a clearer form. The relatively low importance of the number of the embedding parameters for the final performance has been noted by \citet{lan2020albert}.

\newpage
\section{Exploring the Optimal Mamba to MoE Active Parameters Ratio} \label{app:optimal_ratio}
The assignment of FLOPs and parameters to different components is an important design choice in heterogeneous architectures. For example, in Transformer, the shape of the model has been studied extensively by \citet{kaplan2020scaling}. In our work, we investigate the optimal ratio of active parameters in the Mamba layer to the number of active parameters in the MoE layer, see section \ref{sec:optimal_ratio}. Figure \ref{fig:optimal_ratio} may suggest that increasing the ratio strengthens the performance and maybe assigning all the active parameters to Mamba would result in the best performance (ratio \textit{``6:0''}). 
It should, however, be noted, that all the investigated models contain the same number of both total parameters and active parameters per token. A hypothetical model described above could not achieve this property. If we loosen the requirement and place all the parameters in Mamba, the resulting model is the same as \mambaSmall{} with the expansion factor $E=4$ and $8$ instead of $16$ Mamba layers. This model achieves marginally worse final log perplexity than \mambaSmall{} (3.73).

\section{Train and Test Set Performance} \label{app:test_loss}
\begin{figure}[ht] 
\centering
\includegraphics[width=\textwidth]{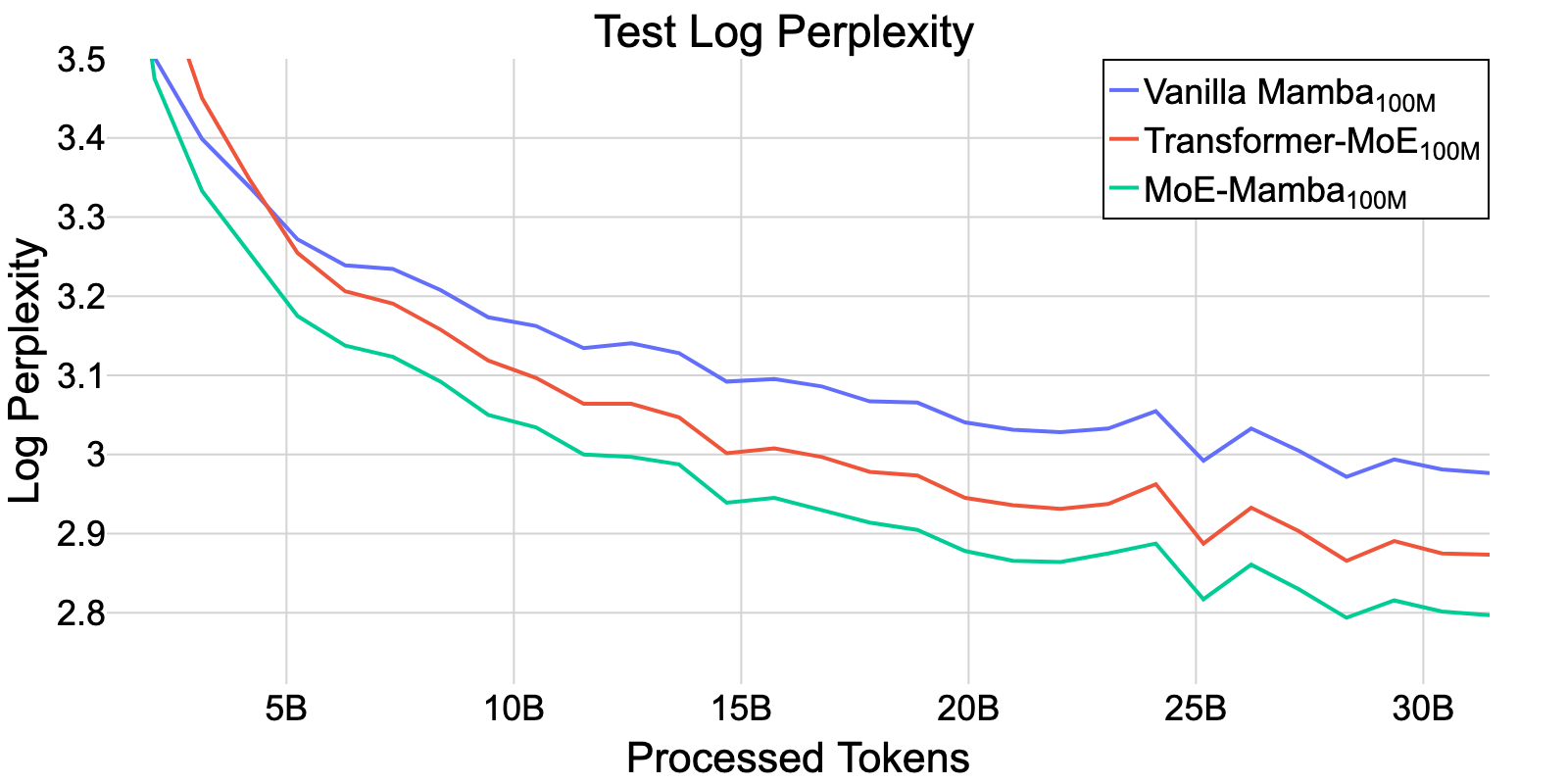}
\caption{Test set loss.} %
\label{app:test_set_performance}
\end{figure}

In the main text, we report the loss values obtained on the train set. Our training procedure samples from the dataset, so even without processing more tokens than there are in the C4 dataset, the same documents may be encountered multiple times. However, as we process less than 20\% of the tokens, the difference in performance on the train set and on the test set should be negligible. For transparency, we provide the results on the test set as well (Figure \ref{app:test_set_performance}). Their variance may be high due to a limited number of sequences in each evaluation step.

\section{Contributions}
\label{app:contributions}
Maciej integrated Mamba into the codebase, ran experiments related to various aspects of this work, and oversaw the course of the project. Kamil ran the bulk of the experiments. Krystian explored alternative Mamba block designs with Jan's help. Michał and Jakub contributed to the project in various ways, mostly by running experiments and perfecting the codebase. Szymon contributed to our shared repo and participated in some early discussions on integrating MoE and SSMs. Piotr and Marek provided high-level scientific advice. Sebastian supervised the project, setting the research direction and leading experiments and analyses.

\section{Reproducibility}
The codebase used to run the experiments is available at \href{https://github.com/llm-random/llm-random}{https://github.com/llm-random/llm-random}.

\end{document}